\documentclass[]{fairmeta}

\usepackage{lineno}
\usepackage{graphicx}
\usepackage{amsmath}
\usepackage{amsfonts}
\usepackage{mathtools}
\usepackage{tcolorbox}
\usepackage{listings}
\usepackage{multirow}
\usepackage{stfloats}
\usepackage{booktabs}
\usepackage{makecell}
\usepackage{pifont}
\usepackage{color-edits} 
\usepackage{cleveref}
\usepackage{makecell}

\newtheorem{definition}{Definition}

\crefname{definition}{Definition}{Definitions}

\title{Imbalanced Gradients in RL Post-Training of Multi-Task LLMs}

\author[1,3,*]{Runzhe Wu}
\author[1,4]{Ankur Samanta}
\author[1]{Ayush Jain}
\author[2]{Scott Fujimoto}
\author[1]{Jeongyeol Kwon}
\author[5]{Ben Kretzu}
\author[1]{Youliang Yu}
\author[2]{Kaveh Hassani}
\author[1]{Boris Vidolov}
\author[1,6,*]{Yonathan Efroni}

\affiliation[1]{Meta AI}
\affiliation[2]{Meta Superintelligence Labs}
\affiliation[3]{Cornell Tech}
\affiliation[4]{Columbia University}
\affiliation[5]{Technion}
\affiliation[6]{Tel Aviv University}

\contribution[*]{Work done at Meta}

\abstract{
    Multi-task post-training of large language models (LLMs) is typically performed by mixing datasets from different tasks and optimizing them jointly. This approach implicitly assumes that all tasks contribute gradients of similar magnitudes; when this assumption fails, optimization becomes biased toward large-gradient tasks. 
    In this paper, however, we show that this assumption fails in RL post-training: certain tasks produce significantly larger gradients, thus biasing updates toward those tasks. Such gradient imbalance would be justified only if larger gradients implied larger learning gains on the tasks (i.e., larger performance improvements)---but we find this is not true. Large-gradient tasks can achieve similar or even much lower learning gains than small-gradient ones. Further analyses reveal that these gradient imbalances cannot be explained by typical training statistics such as training rewards or advantages, suggesting that they arise from the \emph{inherent} differences between tasks. This cautions against naive dataset mixing and calls for future work on principled gradient-level corrections for LLMs.}

\date{\today}
\correspondence{Runzhe Wu at \email{rw646@cornell.edu}}

\begin{document}

\maketitle

\section{Introduction}

Large language models (LLMs) are increasingly trained to master multiple tasks simultaneously, from language understanding and summarization to code generation and math problem solving. Such multi-task learning enables models to generalize across domains and learn more efficiently than by training separate models per task~\citep{qi2024optimizing,brief2024mixing}. 

A standard approach to multi-task post-training is to merge datasets from different tasks and train on them jointly~\citep{mosaicml2023introducing,wang2023aurora,wang2023far,hu2024minicpm,zhu2025dynamic}. While simple, this strategy effectively mixes gradients across tasks and implicitly requires gradient magnitudes to be similar across tasks. When some tasks generate much larger gradients, the optimization becomes dominated by them. Such imbalance has been noted in the broader field such as vision~\citep{chen2018gradnorm} but is less explored in the LLM literature (\Cref{sec:works}).

In this paper, we show that such gradient imbalance arises prominently in the RL post-training of multi-task LLMs (\Cref{fig:intro}). The consequence of this imbalance is detrimental: (1) when certain tasks produce much larger gradients, the gradients aggregated from all tasks are dominated by them, which biases updates toward large-gradient tasks and reduces progress on the rest; and (2) from an optimization perspective, large-gradient tasks are effectively trained as if their learning rates were set to be larger than those of small-gradient tasks. Therefore, either large-gradient tasks are optimized too aggressively, or small-gradient tasks are under-optimized.

Crucially, such gradient imbalances would only be reasonable if larger gradients correspond to greater learning gains; i.e., large-gradient tasks should dominate optimization only when they indeed yield larger improvements on their respective objectives (in our case, training rewards). However, we find this hypothesis does not hold: tasks with much larger gradients can exhibit \emph{similar or even lower} learning gains than those with much smaller gradients.
To further examine the correlation between gradient magnitude and learning gains, we implement gradient-proportional sampling that allocates more training to large-gradient tasks to see if it brings any advantage---the rationale is that if gradient magnitude truly reflects learning gains, then training more frequently on large-gradient tasks should improve overall average performance by prioritizing tasks with greater gains while sacrificing those with smaller ones~\citep{chen2025self,wang2025dump}.
However, we found this approach shows no advantages and can perform even worse in our experiments, proving that the gradient signal across tasks is not merely uncorrelated with learning gains but misleading when used for training signals.
In summary, we find that gradient imbalance is not caused by imbalanced learning gains and can thus harm multi-task learning by biasing updates.

To probe the source of this imbalance, we examined some training statistics as potential explanations, such as training rewards and advantage functions, but we found none could account for it. It suggests that the gradient imbalance does not depend on these training statistics but arises from inherent differences between tasks. This highlights the need for future methods that explicitly correct it in training.

\section{Imbalanced Gradients}\label{sec:imb}

\begin{figure*}[t] 
\includegraphics[width=\linewidth]{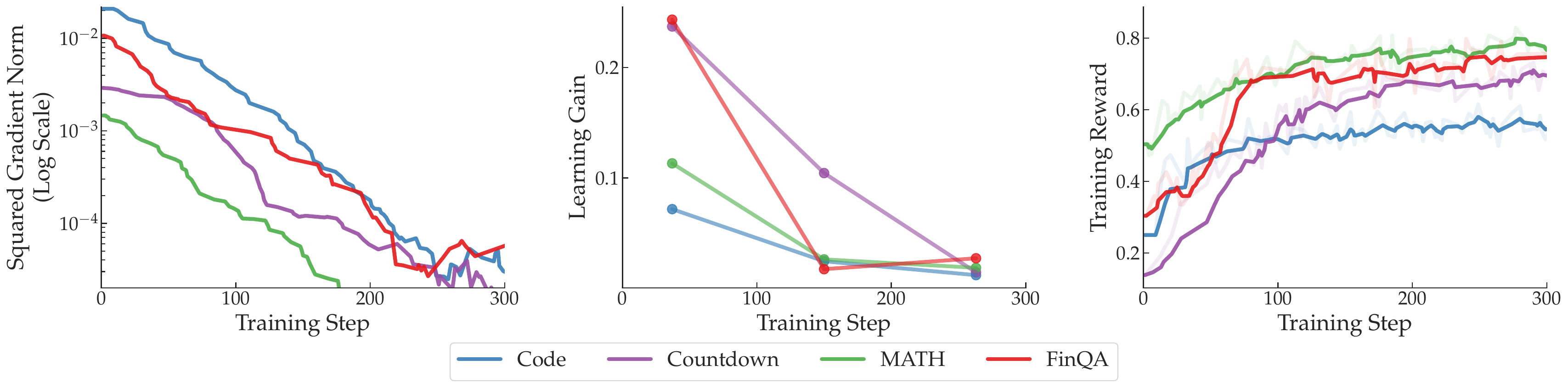}

\vspace{0.7em}

\includegraphics[width=\linewidth]{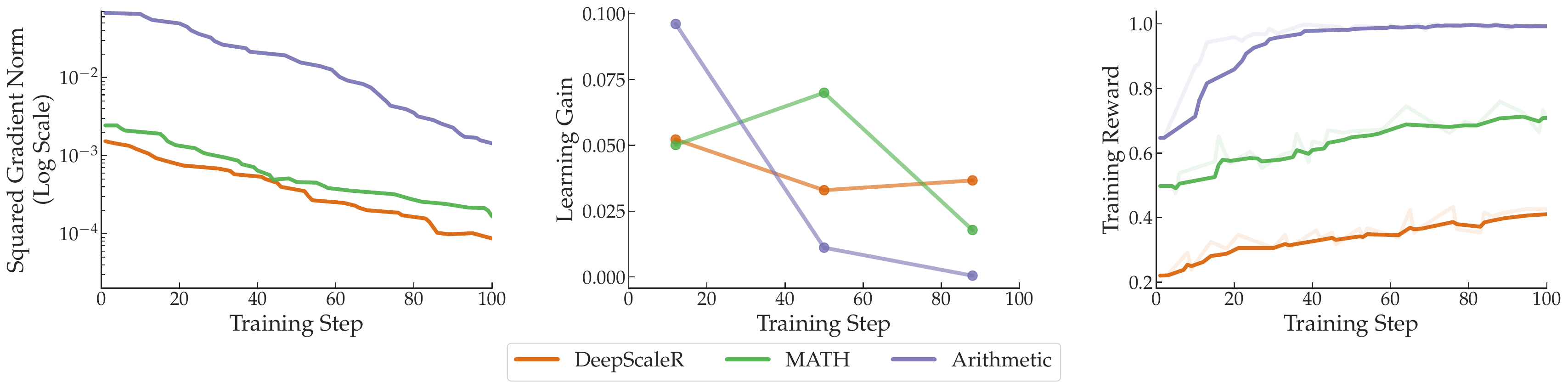}

\vspace{-0.3em}

\caption{ 
    \textbf{Gradient imbalance and its misalignment with learning gains.} 
    \textit{Top:} Qwen-7B trained on four heterogeneous tasks. The left panel shows squared gradient norms, with Code dominating throughout training. However, the middle panel reveals that Code achieves the \emph{lowest} learning gains---defined as the increase in training reward (\Cref{sec:quantifying-learning-gains}). The right panel reports training rewards for reference. 
    \textit{Bottom:} Qwen-7B trained on three math tasks. Arithmetic exhibits up to $33\times$ larger squared gradients but major learning gains vanish after early training. 
    Together, these results indicate an implicit training bias due to imbalanced gradients which cannot be attributed to learning gains. Plots of other models are provided in \Cref{sec:additional-plots}.
}

\label{fig:intro}
\end{figure*}

In this section, we provide the empirical evidence of gradient imbalance in multi-task RL post-training. We study different models: instruction-tuned versions of Qwen2.5-3B and 7B~\citep{qwen2024qwen2} and Llama-3.2-3B~\citep{dubey2024llama}. We consider RLVR~\citep{guo2025deepseek} and use GRPO~\citep{shao2024deepseekmath} as the RL algorithm. 
Additional details of the implementation and the results are reported in \Cref{sec:exp-details,sec:additional-exp-results}. We consider two multi-task settings: (1) four different tasks across multiple domains and (2) three tasks within the math domain but with varying difficulty. They are described below.

\paragraph{Multi-domain tasks.} We focus on four tasks:  
(1) \emph{Code}, involving code generation in the style of programming contests~\citep{code-r1};  
(2) \emph{Countdown}, where the model needs to construct a target number from given inputs using basic arithmetic operators~\citep{tinyzero};  
(3) \emph{MATH}, focused on mathematical problem solving~\citep{hendrycksmath2021}; and  
(4) \emph{FinQA}, financial reasoning over tabular data and text~\citep{chen2021finqa}. 
This setup focuses on examining the gradient behaviors across heterogeneous tasks.

\paragraph{Single-domain (math) tasks.} We consider three datasets related to mathematics. \emph{DeepScaleR}~\citep{deepscaler2025} and \emph{MATH}~\citep{hendrycksmath2021} contain challenging math problems (with DeepScaleR being harder overall). \emph{Arithmetic}~\citep{brown2020language} consists of basic arithmetic problems and is easier. This setup allows us to examine gradient behavior within a single domain while varying task difficulty and problem styles.

We conduct experiments for both multi-task settings described above. During training, batches are constructed by sampling from all tasks with equal probability---importantly, we sample with equal probability rather than simply mixing the datasets from all tasks because the latter leads to a natural bias due to different dataset sizes across tasks.
To study gradient imbalance, we track the average squared gradient norms of each task throughout training and report them in \Cref{fig:intro}. 
We focus on \emph{squared} norms rather than unsquared ones because theoretically they are more directly connected to learning gains (\Cref{sec:convex-analysis}), which we will compare against in subsequent sections. For completeness, we also report the unsquared norm in \Cref{sec:additional-plots} (\Cref{fig:unsq_grad_norm}) where we observe the identical trend.

From \Cref{fig:intro} (left column), we observe that the gradient magnitudes are decreasing as training proceeds for all tasks. However, there is a clear dominance pattern across tasks. For instance, in the multi-domain setting (top left), the gradient of Code has been the largest throughout training---its squared norm is up to $15\times$ larger than MATH; in the single-domain setting (bottom left), Arithmetic has been dominating the others with squared norms up to $33\times$ larger. Such an issue is severe if we consider how gradients are averaged during training. Suppose we have $M$ tasks, each with gradient $g_i$ for $i\in[M]$. A uniform mixture leads to the average gradient $\bar{g} = \frac{1}{M} \sum_{i \in [M]} g_i$.
If certain tasks have substantially larger gradients than others 
(e.g., Arithmetic exhibiting significantly larger gradients than the others), the averaged gradient becomes effectively dominated by those tasks: $\bar{g} \approx \frac{1}{M} \sum_{i \in \mathcal{I}} g_i$
where $\mathcal{I}$ denotes the set of tasks with outstandingly large gradients. 
As a result, optimization is biased toward these tasks while under-optimizing the rest ($[M]\setminus \mathcal{I}$). Another way to view this effect is through the lens of the learning rate: with a globally fixed learning rate, a task such as Arithmetic (with gradient norms up to $\sqrt{33}\approx5.7\times$ larger than others) was effectively trained \emph{as if its learning rate were $5.7\times$ higher than others}. Hence, to stabilize training, the global learning rate must then be reduced, which in turn causes small-gradient tasks to be under-optimized. This runs counter to the goal of multi-task learning, where all tasks should be treated equally.

\section{Are Gradient Imbalances Explained by Learning Gains?}\label{sec:learning-gains}

Having established the presence of gradient imbalances, we next ask whether they can be explained by learning gains. The reasoning is simple: in multi-task training, large gradients can be justified only if they faithfully reflect larger room for improvement than the others. Only in this case will the gradient bias be not harmful and even be desirable. This hypothesis underlies several successful approaches in the broader machine learning literature~\citep{settles2007multiple,ashdeep,chen2025self}. However, whether it holds in multi-task RL post-training remains to be validated. We examine it in two ways: (1) by explicitly quantifying learning gains as the increase in training reward and comparing it with gradient magnitudes, and (2) by studying a gradient-proportional sampling strategy. Our results show that, while gradient magnitude tends to be correlated with learning gains \emph{within a task}, which is consistent with the prior intuition, it does not hold \emph{across tasks}, which is the central issue highlighted in this paper. We elaborate on the two methods in the following two subsections.

\subsection{Quantifying Learning Gains}\label{sec:quantifying-learning-gains}

As a direct approach, we explicitly measure the learning gains as the increase in training reward and compare it with gradient magnitudes to see if they are correlated. Particularly, the learning gain on a given task at training step $t$ is measured as the average change in training reward around $t$:
\[
\text{Gain}(t) 
\coloneqq
\frac{1}{s} \sum_{i=1}^{s} R_{t+i}
-
\frac{1}{s} \sum_{i=1}^{s} R_{t-i},
\]
where $R_{k}$ denotes training reward at step $k$, and $s$ is the window size used to smooth rewards. We evaluate the learning gains at three evenly spaced training steps and plot their corresponding values in \Cref{fig:intro} (middle column). It reveals a different pattern compared to gradient magnitudes (left column). In particular, in the multi-domain setting (top), Code dominates all others in terms of gradient magnitude, yet its gains are one of the \emph{lowest}. The contrast is clearer in the single-domain setting (bottom): Arithmetic constantly exhibits up to $33\times$ larger squared gradient norms than the others, but it shows the lowest learning gains after half of training. Hence, differences in gradient magnitude across tasks cannot be attributed to differences in learning gains. 

\subsection{Gradient-Proportional Sampling}\label{sec:gradient-proportional-sampling}

\begin{table*}[t]
\small
\centering
\begin{NiceTabular}{lcccccccccc}
    \toprule
    \multirow{2}{*}{{Model}} 
    & \multirow{2}{*}{\makecell{Grad-Prop\\Sampling?}}
    & \multicolumn{5}{c}{{Multi-domain}} 
    & \multicolumn{4}{c}{{Single-domain (math)}} 
    \\
    \cmidrule(lr{2pt}){3-7}
    \cmidrule(l{2pt}r){8-11}
    & & {Code} & {Countdown} & {MATH} & {FinQA} & \textit{{Avg}} & {DSR} & {MATH} & {Arith} & \textit{{Avg}}
    \\
    \midrule
    \multirow{2}{*}{Qwen-3B} & \ding{55} & 35.67 & \textbf{50.00} & \textbf{63.20} & 60.60 & \textbf{52.37} & \textbf{34.00} & 58.00 & 98.00 & 63.33 \\
    & \ding{51} & 35.67 & 42.50 & 61.00 & \textbf{62.67} & 50.46 & 33.40 & \textbf{61.20} & 98.00 & \textbf{64.20} \\
    \midrule
    \multirow{2}{*}{Lllama-3B} & \ding{55} & 30.34 & \textbf{42.50} & \textbf{49.00} & 64.65 & \textbf{46.62} & \textbf{26.20} & 45.00 & 95.60 & \textbf{55.60} \\
    & \ding{51} & \textbf{30.76} & 40.50 & 48.20 & \textbf{65.01} & 46.12 & 23.60 & \textbf{46.00} & \textbf{96.00} & 55.20 \\
    \midrule
    \multirow{2}{*}{Qwen-7B} & \ding{55} & 49.16 & \textbf{59.00} & \textbf{71.00} & \textbf{68.71} & \textbf{61.97} & \textbf{41.40} & \textbf{68.40} & 99.20 & \textbf{69.67} \\
    & \ding{51} & \textbf{50.70} & 53.00 & 69.60 & 68.62 & 60.48 & 34.20 & 64.20 & \textbf{99.60} & 66.00 \\
    \bottomrule
\end{NiceTabular}
\caption{
    \textbf{Uniform vs. gradient-proportional sampling.} 
    Across models and multi-task settings, sampling tasks in proportion to gradient magnitude raises accuracy on some large-gradient tasks (e.g., FinQA, Arithmetic) but can lower overall averages. This confirms that large gradients do not necessarily correspond to greater learning gains, so prioritizing them hurts multi-task learning. 
    (DSR: DeepScaleR, Arith: Arithmetic, Avg: Average)
}
\label{tab:4t}
\end{table*}

Another way to examine the hypothesis that the gradient magnitude is positively correlated with larger learning gains is to test algorithms that explicitly exploit large gradient magnitudes. Hence, we study a gradient-proportional sampling strategy: to form the training batches at each iteration, instead of sampling tasks uniformly, we sample from tasks in proportion to their average gradient magnitudes. Specifically, given $M$ tasks, we define the probability of sampling from task $i\in[M]$ as the softmax of the gradient magnitudes: $p_i = \exp(\|g_i\| / \eta) / \sum_j \exp(\|g_j\| / \eta)$ where $g_i$ denotes the average gradient of task $i$ and $\eta$ is the temperature. 
To avoid the model getting stuck on high-gradient tasks under low-temperature settings, we truncate $p_i$ at $0.1$, ensuring that every task is sampled with at least a $10\%$ probability. 
The rationale behind gradient-proportional sampling is straightforward: if gradient magnitude truly reflects learning gains, then biasing training toward large-gradient tasks should improve overall average performance by prioritizing tasks with greater gains while sacrificing those with smaller ones. 

We report the comparison between gradient-proportional sampling and uniform sampling on both multi-task settings across various models in \Cref{tab:4t}. The results show no advantage of this approach. While it primarily raises test accuracy on the large-gradient tasks as expected (e.g., Code, FinQA and Arithmetic, evidenced in \Cref{fig:intro}), it can lead to larger drops on other tasks, thus leading to lower average performance.
This again suggests that gradient magnitude is not a reliable indicator of learning gains: allocating more training to large-gradient tasks does not consistently improve overall outcomes and may even exacerbate gradient bias.
Importantly, our intention is not to claim that gradient-proportional sampling is consistently worse than uniform sampling; if that were the case, one might conclude gradient magnitude is inversely correlated with learning gains, which is counterintuitive and unlikely. 
Rather, we aim to show that gradient magnitude cannot faithfully represent learning gains.
Detailed training curves and ablations of gradient-proportional sampling are provided in \Cref{sec:ablation-study,sec:additional-plots} for reference.

\paragraph{Summary.}
Through explicit measurement~(\Cref{sec:quantifying-learning-gains}) and implicit testing~(\Cref{sec:gradient-proportional-sampling}), we show gradient magnitude does not correlate with learning gains across tasks. We also provide a theoretical analysis in \Cref{sec:convex-analysis} explaining why learning gains do not necessarily correlate with gradient magnitude across tasks from the perspective of convex analysis.

\section{Are Gradient Imbalances Explained by Other Training Statistics?}\label{sec:other-training-statistics}

In this section, we study whether gradient magnitude correlates with several key training statistics and metrics, including advantage function, training reward, and token length. We note that, even if such correlations exist, gradient imbalance would remain problematic because it cannot reflect learning gains, as we discussed in \Cref{sec:learning-gains}.

\paragraph{Advantage Function.} 
\begin{figure*}[t] 
\centering
\includegraphics[width=0.32\linewidth]{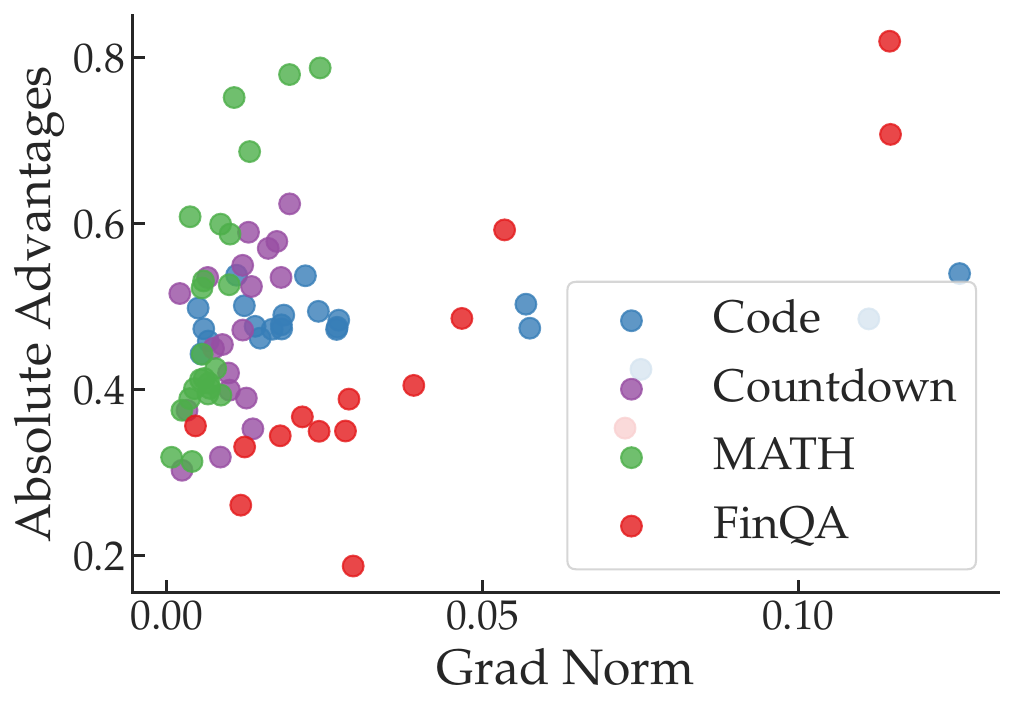}
\hfill
\includegraphics[width=0.32\linewidth]{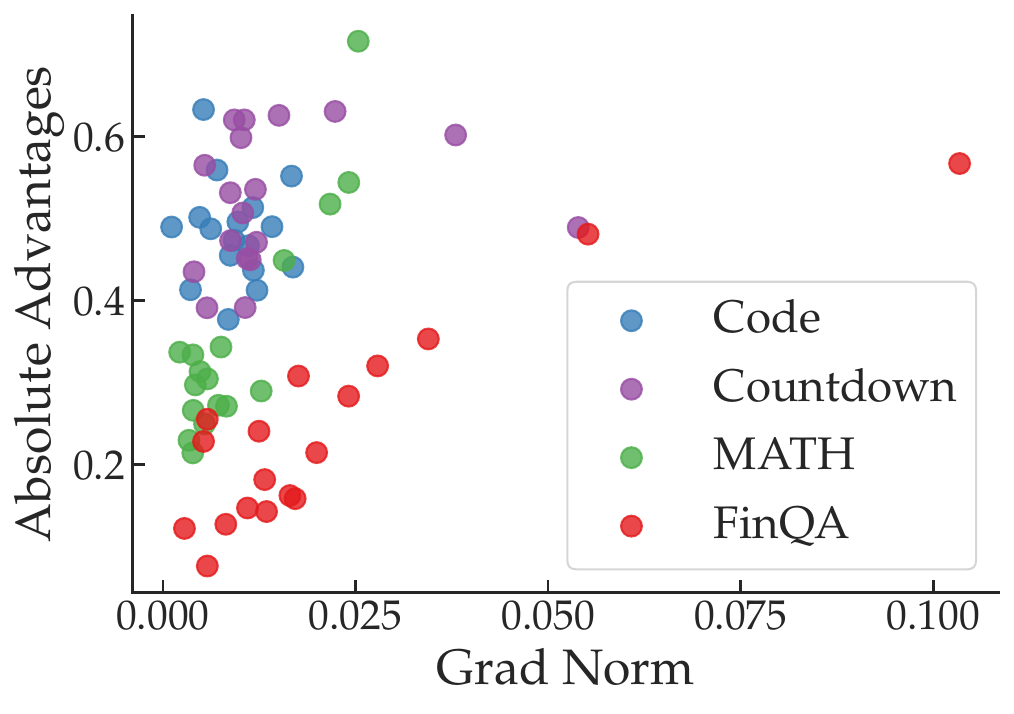}
\hfill
\includegraphics[width=0.32\linewidth]{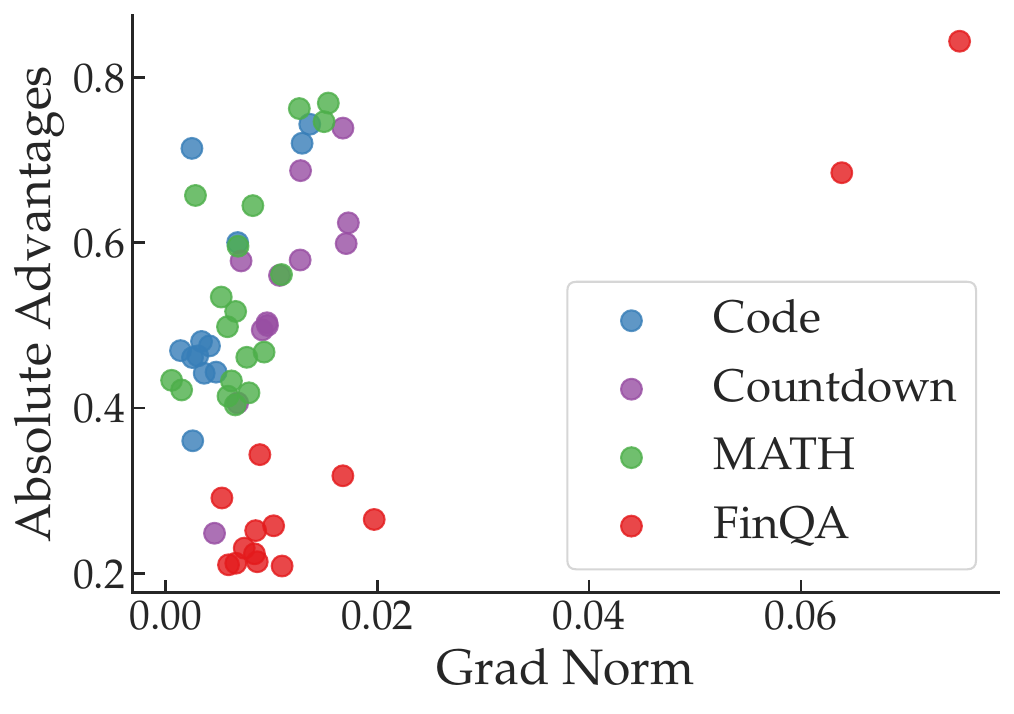}
\includegraphics[width=0.32\linewidth]{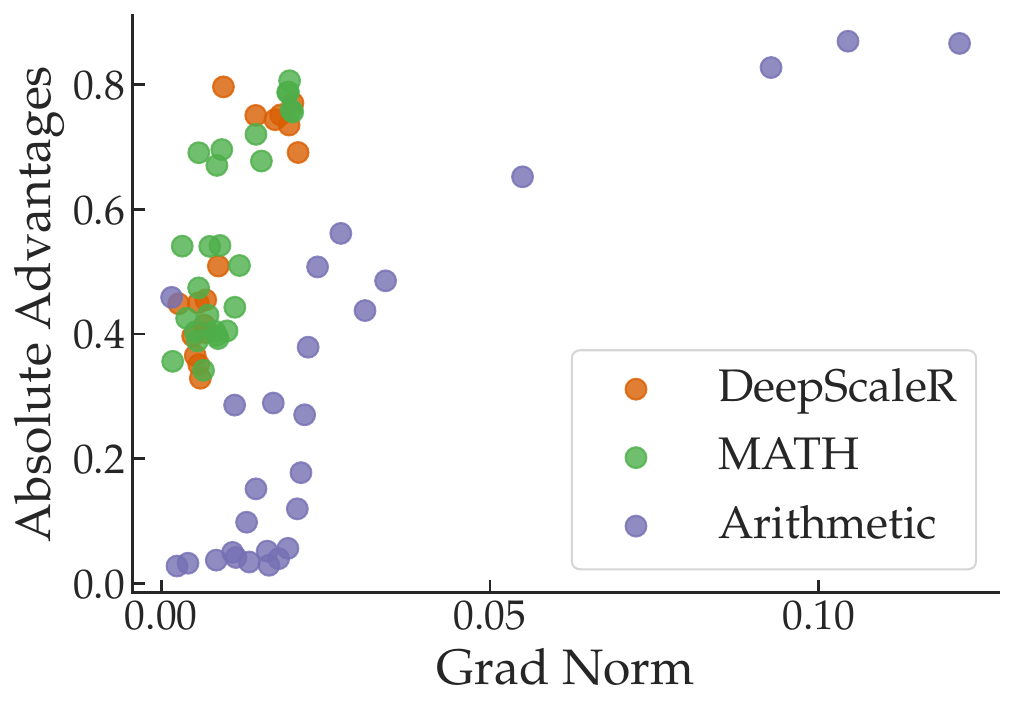}
\hfill
\includegraphics[width=0.32\linewidth]{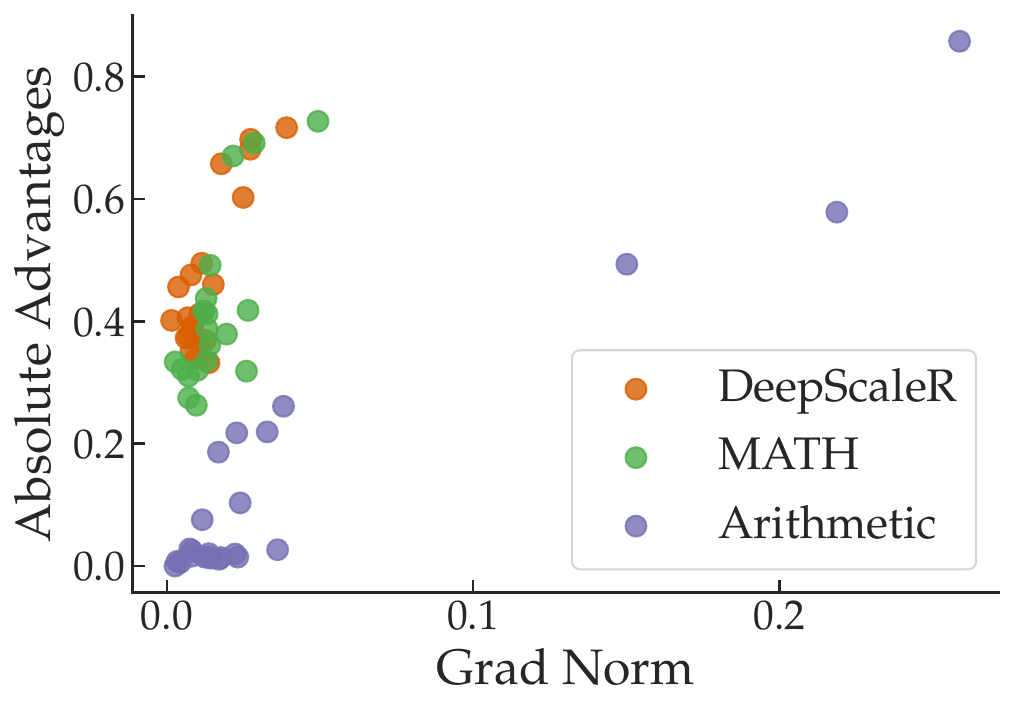}
\hfill
\includegraphics[width=0.32\linewidth]{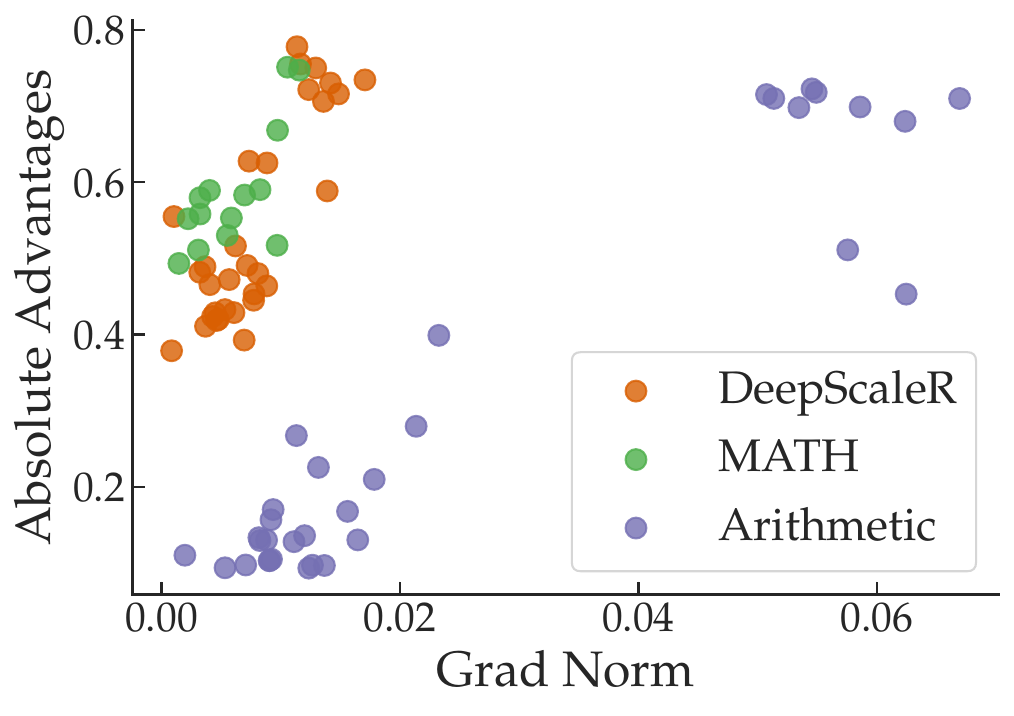}
\caption{
    \textbf{Gradient norms versus absolute advantage.} 
    Scatter plots across models (from left to right: Qwen-3B, Qwen-7B, Llama-3B) and domains (from top to down: multi-domain and math-domain). 
    Within each task, points tend to cluster and vary consistently. Within a task, larger gradients often align with larger absolute advantages, although unnecessarily linearly. 
    Across tasks, however, this structure disappears: tasks with the largest gradients (e.g., FinQA) do not yield higher absolute advantages than other tasks. 
    Thus, absolute advantage may approximate gradient magnitude \emph{within} a task but fails \emph{across} tasks.
}

\label{fig:adv_vs_grad_norm}
\end{figure*}
Prior work has speculated that the absolute advantage function can serve as an effective proxy for gradient magnitude of language models, mainly focusing on single-task training~\citep{chen2025self,wang2025dump,gao2025prompt}. To see the intuition, we recall the policy-gradient theorem. For policy-gradient-based RL, the derivative of the model parameters is derived as~\citep{sutton1998reinforcement}
\begin{align*}
\nabla_{\theta} J(\theta) = \mathbb{E}[ \nabla_\theta \log \pi_\theta (a \,|\, s) A^\pi(s,a)]
\end{align*}
where $J(\theta)$ denotes the average reward of the policy $\pi_\theta$, and $A^\pi(s,a)$ is the advantage function. By Jensen's inequality, we have
\begin{align*}
&\|\nabla_{\theta} J(\theta)\|_2 \leq \mathbb{E} \big[ |A^\pi(s,a)|\|\nabla_\theta \log \pi_\theta (a \,|\, s)\|_2 \big]
\end{align*}
Hence, $|A^\pi|$ seems to be an effective proxy for gradient magnitude, assuming the inequality is nearly tight and $\nabla_\theta \log \pi_\theta$ does not vary too much. 
While this hypothesis has led to successful approaches in single-task training, it remains to be seen whether it holds in multi-task training. Thus, we put it to the test and report results in \Cref{fig:adv_vs_grad_norm}, where in the scatter plot, we track and plot the absolute advantage and gradient norm periodically throughout training.
The computation of advantage is following standard GRPO, which is estimated via empirical mean and then normalized. 
Interestingly, we find a peculiar pattern: the absolute advantage is a reasonable proxy for gradient magnitude \emph{within a task}, but it fails to generalize \emph{across tasks}. In particular, within each task, the absolute advantage grows roughly proportionally with the gradient magnitude. However, when comparing across tasks, the same level of absolute advantage corresponds to very different gradient norms: for example, MATH tends to yield much smaller gradients than FinQA. Hence, we conclude that the relationship is valid only within a single task.

\paragraph{Training Reward (Accuracy).}
It is also tempting to associate gradient magnitude with task difficulty. For example, we may speculate that easy tasks (high average reward/accuracy) yield clearer, larger gradients while hard tasks (low average reward/accuracy) produce noisier, smaller gradients on average. However, this intuition may not hold in the extreme: when accuracy is near \(0\%\) or \(100\%\), the advantage \(A^\pi\) is close to zero; by the policy-gradient theorem the expected gradient also approaches zero~\citep{yu2025dapo,zhang2025speed}. Thus, gradient magnitude need not increase monotonically with ``easiness'' or ``hardness''. Qualitatively, in \Cref{fig:intro} (right columns), while within each task, the trend shows gradient norms decreasing as training reward (accuracy) increases, there is no consistent cross-task pattern, suggesting that there is unlikely to be a simple relationship between gradient magnitude and task difficulty.
\begin{figure}[t]
\centering
\begin{minipage}[c]{0.35\linewidth}
\small
\centering
\begin{tabular}{llc}
    \toprule
    & {Task} & {Prompt Len} \\
    \midrule
    \multirow{4}{*}{\makecell{{Multi-}\\{domain}}}
    & Code & 553.2 \\
    & Countdown & 164.5 \\
    & MATH & 168.6 \\
    & FinQA & 1161.9 \\
    \midrule
    \multirow{3}{*}{\makecell{{Single-}\\{domain}}}
    & DeepScaleR & 173.1 \\
    & MATH & 168.6 \\
    & Arithmetic & 105.4 \\
    \bottomrule
\end{tabular}
\end{minipage}
\hfill
\begin{minipage}[c]{0.64\linewidth}
\centering
\includegraphics[width=0.49\linewidth]{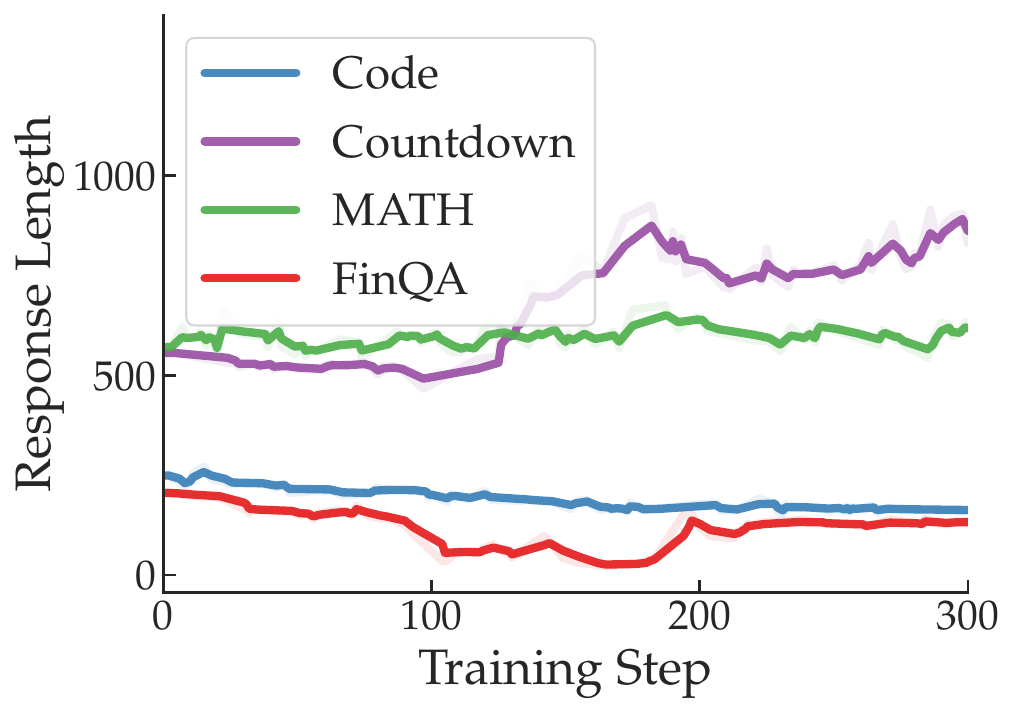}
\includegraphics[width=0.49\linewidth]{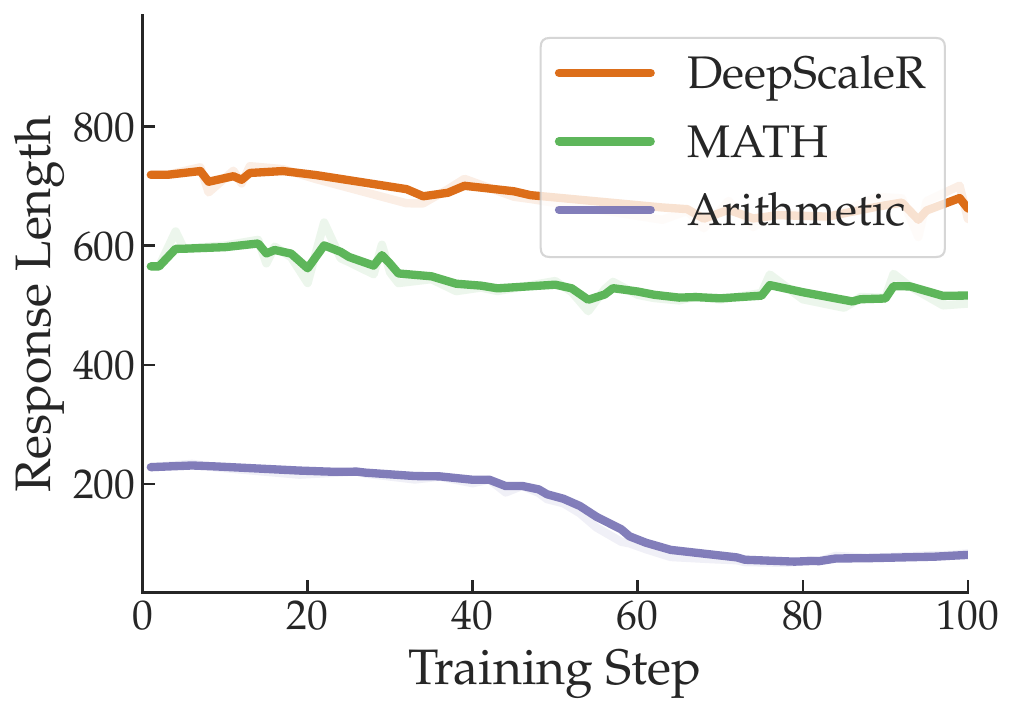}
\end{minipage}
\caption{\textbf{Prompt and response lengths across tasks.} 
Left: Average prompt length for each task. 
Middle and right: Curves of response length for Qwen-3B model through training, for multi-domain and single-domain tasks respectively.}
\label{tab:prompt-length}
\label{fig:response-length}
\end{figure}

\paragraph{Prompt/Response Length.}
Since different tasks have varying prompt and response lengths, we also investigate whether gradient magnitude is affected by this factor. In principle, it should not be, because the policy-gradient loss is averaged over tokens and is therefore expected to be invariant to sequence length. Nonetheless, we examine the relationship empirically. The average prompt lengths for all tasks are reported in \Cref{tab:prompt-length}. We find no meaningful correlation between prompt length and gradient magnitude. FinQA often has long prompts (due to embedded table data) and tends to exhibit large gradients, yet this pattern does not hold universally: in the single-domain setting, Arithmetic has the largest gradient magnitude despite having the shortest average prompt. Similarly, the curves of response length shown in \Cref{fig:response-length} reveal no meaningful connection either.

In summary, we found that the gradient imbalance cannot be easily explained by any of the above training statistics. Hence, the imbalance is likely from the inherent differences between tasks.

\section{Conclusion}

We presented the first systematic study of gradient imbalance in the RL post-training phase of multi-task LLMs. Results show a surprising finding: the gradient imbalance is significant and cannot be explained by learning gains. Further, we study whether the gradient imbalances can be explained by certain training statistics or metrics (e.g., advantages, training reward, token length, etc.) but we find no clear correlation. Hence, the observed gradient imbalance likely stems from inherent differences between tasks.

Looking ahead, our findings call for approaches that explicitly perform gradient-level manipulation to address the imbalances, possibly building on ideas from the broader optimization literature~\citep{sener2018multi,yu2020gradient,chen2020just,liu2021conflict,liutowards,liu2023famo,efronialigned, kretzu2025simple}. Beyond that, another promising direction lies in reconsidering the optimization geometry. Methods such as mirror descent-based RL approaches~\citep{kakade2001natural,gao2024rebel} may help alleviate this issue by transforming gradients into a certain representation that is more comparable across tasks. 
Finally, investigating whether the same issue occurs in other training paradigms such as pre-training and supervised finetuning remains important for future work.

\section*{Limitations}

This paper focuses primarily on the RL post-training of multi-task LLMs. It remains unclear whether gradient imbalance also persists in other phases, such as pre-training and supervised fine-tuning. Although such imbalances appear to be from the inherent differences between tasks, we cannot yet explain why it arises, despite studying a broad range of training statistics. Moreover, we only report these observations; how to design new approaches to mitigate them remains unclear and requires further research. Finally, we do not anticipate any immediate ethical or societal risk from this work, though broader concerns may emerge from the use of LLMs more generally.

\bibliographystyle{assets/plainnat}
\bibliography{paper}

\begin{thebibliography}{49}
\providecommand{\natexlab}[1]{#1}
\providecommand{\url}[1]{\texttt{#1}}
\expandafter\ifx\csname urlstyle\endcsname\relax
  \providecommand{\doi}[1]{doi: #1}\else
  \providecommand{\doi}{doi: \begingroup \urlstyle{rm}\Url}\fi

\bibitem[Ash et~al.(2020)Ash, Zhang, Krishnamurthy, Langford, and Agarwal]{ashdeep}
Jordan~T Ash, Chicheng Zhang, Akshay Krishnamurthy, John Langford, and Alekh Agarwal.
\newblock Deep batch active learning by diverse, uncertain gradient lower bounds.
\newblock In \emph{International Conference on Learning Representations}, 2020.

\bibitem[Brief et~al.(2024)Brief, Ovadia, Shenderovitz, Yoash, Lemberg, and Sheetrit]{brief2024mixing}
Meni Brief, Oded Ovadia, Gil Shenderovitz, Noga~Ben Yoash, Rachel Lemberg, and Eitam Sheetrit.
\newblock Mixing it up: The cocktail effect of multi-task fine-tuning on llm performance--a case study in finance.
\newblock \emph{arXiv preprint arXiv:2410.01109}, 2024.

\bibitem[Brown et~al.(2020)Brown, Mann, Ryder, Subbiah, Kaplan, Dhariwal, Neelakantan, Shyam, Sastry, Askell, Agarwal, Herbert-Voss, Krueger, Henighan, Child, Ramesh, Ziegler, Wu, Winter, Hesse, Chen, Sigler, Litwin, Gray, Chess, Clark, Berner, McCandlish, Radford, Sutskever, and Amodei]{brown2020language}
Tom~B. Brown, Benjamin Mann, Nick Ryder, Melanie Subbiah, Jared Kaplan, Prafulla Dhariwal, Arvind Neelakantan, Pranav Shyam, Girish Sastry, Amanda Askell, Sandhini Agarwal, Ariel Herbert-Voss, Gretchen Krueger, Tom Henighan, Rewon Child, Aditya Ramesh, Daniel~M. Ziegler, Jeffrey Wu, Clemens Winter, Christopher Hesse, Mark Chen, Eric Sigler, Mateusz Litwin, Scott Gray, Benjamin Chess, Jack Clark, Christopher Berner, Sam McCandlish, Alec Radford, Ilya Sutskever, and Dario Amodei.
\newblock Language models are few-shot learners.
\newblock 2020.

\bibitem[Chen et~al.(2025)Chen, Lu, Kim, Zhang, Tang, Pich{\'e}, Gontier, Bengio, and Kamalloo]{chen2025self}
Xiaoyin Chen, Jiarui Lu, Minsu Kim, Dinghuai Zhang, Jian Tang, Alexandre Pich{\'e}, Nicolas Gontier, Yoshua Bengio, and Ehsan Kamalloo.
\newblock Self-evolving curriculum for llm reasoning.
\newblock \emph{arXiv preprint arXiv:2505.14970}, 2025.

\bibitem[Chen et~al.(2018)Chen, Badrinarayanan, Lee, and Rabinovich]{chen2018gradnorm}
Zhao Chen, Vijay Badrinarayanan, Chen-Yu Lee, and Andrew Rabinovich.
\newblock Gradnorm: Gradient normalization for adaptive loss balancing in deep multitask networks.
\newblock In \emph{International conference on machine learning}, pages 794--803. PMLR, 2018.

\bibitem[Chen et~al.(2020)Chen, Ngiam, Huang, Luong, Kretzschmar, Chai, and Anguelov]{chen2020just}
Zhao Chen, Jiquan Ngiam, Yanping Huang, Thang Luong, Henrik Kretzschmar, Yuning Chai, and Dragomir Anguelov.
\newblock Just pick a sign: Optimizing deep multitask models with gradient sign dropout.
\newblock \emph{Advances in Neural Information Processing Systems}, 33:\penalty0 2039--2050, 2020.

\bibitem[Chen et~al.(2021)Chen, Chen, Smiley, Shah, Borova, Langdon, Moussa, Beane, Huang, Routledge, et~al.]{chen2021finqa}
Zhiyu Chen, Wenhu Chen, Charese Smiley, Sameena Shah, Iana Borova, Dylan Langdon, Reema Moussa, Matt Beane, Ting-Hao Huang, Bryan~R Routledge, et~al.
\newblock Finqa: A dataset of numerical reasoning over financial data.
\newblock In \emph{Proceedings of the 2021 Conference on Empirical Methods in Natural Language Processing}, pages 3697--3711, 2021.

\bibitem[Dubey et~al.(2024)Dubey, Jauhri, Pandey, Kadian, Al-Dahle, Letman, Mathur, Schelten, Yang, Fan, et~al.]{dubey2024llama}
Abhimanyu Dubey, Abhinav Jauhri, Abhinav Pandey, Abhishek Kadian, Ahmad Al-Dahle, Aiesha Letman, Akhil Mathur, Alan Schelten, Amy Yang, Angela Fan, et~al.
\newblock The llama 3 herd of models.
\newblock \emph{arXiv e-prints}, pages arXiv--2407, 2024.

\bibitem[Efroni et~al.(2025)Efroni, Kretzu, Jiang, Bhandari, Zhu, and Ullrich]{efronialigned}
Yonathan Efroni, Ben Kretzu, Daniel Jiang, Jalaj Bhandari, Zheqing Zhu, and Karen Ullrich.
\newblock Aligned multi objective optimization.
\newblock In \emph{Forty-second International Conference on Machine Learning}, 2025.

\bibitem[Elbakary et~al.(2025)Elbakary, Issaid, ElBatt, Seddik, and Bennis]{elbakary2025mira}
Ahmed Elbakary, Chaouki~Ben Issaid, Tamer ElBatt, Karim Seddik, and Mehdi Bennis.
\newblock Mira: A method of federated multi-task learning for large language models.
\newblock \emph{IEEE Networking Letters}, 2025.

\bibitem[Feng et~al.(2024)Feng, Hao, Zhang, Han, and Wang]{feng2024mixture}
Wenfeng Feng, Chuzhan Hao, Yuewei Zhang, Yu~Han, and Hao Wang.
\newblock Mixture-of-loras: An efficient multitask tuning method for large language models.
\newblock In \emph{Proceedings of the 2024 Joint International Conference on Computational Linguistics, Language Resources and Evaluation (LREC-COLING 2024)}, pages 11371--11380, 2024.

\bibitem[Gao et~al.(2024)Gao, Chang, Zhan, Oertell, Swamy, Brantley, Joachims, Bagnell, Lee, and Sun]{gao2024rebel}
Zhaolin Gao, Jonathan Chang, Wenhao Zhan, Owen Oertell, Gokul Swamy, Kiant{\'e} Brantley, Thorsten Joachims, Drew Bagnell, Jason~D Lee, and Wen Sun.
\newblock Rebel: Reinforcement learning via regressing relative rewards.
\newblock \emph{Advances in Neural Information Processing Systems}, 37:\penalty0 52354--52400, 2024.

\bibitem[Gao et~al.(2025)Gao, Kim, Sun, Joachims, Wang, Pang, and Tan]{gao2025prompt}
Zhaolin Gao, Joongwon Kim, Wen Sun, Thorsten Joachims, Sid Wang, Richard~Yuanzhe Pang, and Liang Tan.
\newblock Prompt curriculum learning for efficient llm post-training.
\newblock \emph{arXiv preprint arXiv:2510.01135}, 2025.

\bibitem[Gong et~al.(2024)Gong, Yu, Liao, Liu, Chen, and Li]{gong2024coba}
Zi~Gong, Hang Yu, Cong Liao, Bingchang Liu, Chaoyu Chen, and Jianguo Li.
\newblock Coba: Convergence balancer for multitask finetuning of large language models.
\newblock In \emph{Proceedings of the 2024 Conference on Empirical Methods in Natural Language Processing}, pages 8063--8077, 2024.

\bibitem[Guo et~al.(2025)Guo, Yang, Zhang, Song, Zhang, Xu, Zhu, Ma, Wang, Bi, et~al.]{guo2025deepseek}
Daya Guo, Dejian Yang, Haowei Zhang, Junxiao Song, Ruoyu Zhang, Runxin Xu, Qihao Zhu, Shirong Ma, Peiyi Wang, Xiao Bi, et~al.
\newblock Deepseek-r1: Incentivizing reasoning capability in llms via reinforcement learning.
\newblock \emph{arXiv preprint arXiv:2501.12948}, 2025.

\bibitem[Hazan et~al.(2016)]{hazan2016introduction}
Elad Hazan et~al.
\newblock Introduction to online convex optimization.
\newblock \emph{Foundations and Trends{\textregistered} in Optimization}, 2\penalty0 (3-4):\penalty0 157--325, 2016.

\bibitem[Hendrycks et~al.(2021)Hendrycks, Burns, Kadavath, Arora, Basart, Tang, Song, and Steinhardt]{hendrycksmath2021}
Dan Hendrycks, Collin Burns, Saurav Kadavath, Akul Arora, Steven Basart, Eric Tang, Dawn Song, and Jacob Steinhardt.
\newblock Measuring mathematical problem solving with the math dataset.
\newblock \emph{NeurIPS}, 2021.

\bibitem[Hu et~al.(2024)Hu, Tu, Han, Cui, He, Zhao, Long, Zheng, Fang, Huang, Zhang, Thai, Wang, Yao, Zhao, Zhou, Cai, Zhai, Ding, Jia, Zeng, dahai li, Liu, and Sun]{hu2024minicpm}
Shengding Hu, Yuge Tu, Xu~Han, Ganqu Cui, Chaoqun He, Weilin Zhao, Xiang Long, Zhi Zheng, Yewei Fang, Yuxiang Huang, Xinrong Zhang, Zhen~Leng Thai, Chongyi Wang, Yuan Yao, Chenyang Zhao, Jie Zhou, Jie Cai, Zhongwu Zhai, Ning Ding, Chao Jia, Guoyang Zeng, dahai li, Zhiyuan Liu, and Maosong Sun.
\newblock Mini{CPM}: Unveiling the potential of small language models with scalable training strategies.
\newblock In \emph{First Conference on Language Modeling}, 2024.
\newblock \url{https://openreview.net/forum?id=3X2L2TFr0f}.

\bibitem[Kakade(2001)]{kakade2001natural}
Sham~M Kakade.
\newblock A natural policy gradient.
\newblock \emph{Advances in neural information processing systems}, 14, 2001.

\bibitem[Kretzu et~al.(2025)Kretzu, Ullrich, and Efroni]{kretzu2025simple}
Ben Kretzu, Karen Ullrich, and Yonathan Efroni.
\newblock Simple optimizers for convex aligned multi-objective optimization.
\newblock \emph{arXiv preprint arXiv:2509.05811}, 2025.

\bibitem[Liang et~al.(2025)Liang, Qiu, Ding, Liu, Tompkin, Xu, Xia, Tu, Shi, and Zhu]{liang2025modomodo}
Yiqing Liang, Jielin Qiu, Wenhao Ding, Zuxin Liu, James Tompkin, Mengdi Xu, Mengzhou Xia, Zhengzhong Tu, Laixi Shi, and Jiacheng Zhu.
\newblock Modomodo: Multi-domain data mixtures for multimodal llm reinforcement learning.
\newblock \emph{arXiv preprint arXiv:2505.24871}, 2025.

\bibitem[Liu et~al.(2021{\natexlab{a}})Liu, Liu, Jin, Stone, and Liu]{liu2021conflict}
Bo~Liu, Xingchao Liu, Xiaojie Jin, Peter Stone, and Qiang Liu.
\newblock Conflict-averse gradient descent for multi-task learning.
\newblock \emph{Advances in Neural Information Processing Systems}, 34:\penalty0 18878--18890, 2021{\natexlab{a}}.

\bibitem[Liu et~al.(2023)Liu, Feng, Stone, and Liu]{liu2023famo}
Bo~Liu, Yihao Feng, Peter Stone, and Qiang Liu.
\newblock Famo: Fast adaptive multitask optimization.
\newblock \emph{Advances in Neural Information Processing Systems}, 36:\penalty0 57226--57243, 2023.

\bibitem[Liu and Zhang(2025)]{code-r1}
Jiawei Liu and Lingming Zhang.
\newblock Code-r1: Reproducing r1 for code with reliable rewards.
\newblock 2025.

\bibitem[Liu et~al.(2021{\natexlab{b}})Liu, Li, Kuang, Xue, Chen, Yang, Liao, and Zhang]{liutowards}
Liyang Liu, Yi~Li, Zhanghui Kuang, Jing-Hao Xue, Yimin Chen, Wenming Yang, Qingmin Liao, and Wayne Zhang.
\newblock Towards impartial multi-task learning.
\newblock In \emph{International Conference on Learning Representations}, 2021{\natexlab{b}}.

\bibitem[Luo et~al.(2025)Luo, Tan, Wong, Shi, Tang, Roongta, Cai, Luo, Li, Popa, and Stoica]{deepscaler2025}
Michael Luo, Sijun Tan, Justin Wong, Xiaoxiang Shi, William~Y. Tang, Manan Roongta, Colin Cai, Jeffrey Luo, Li~Erran Li, Raluca~Ada Popa, and Ion Stoica.
\newblock Deepscaler: Surpassing o1-preview with a 1.5b model by scaling rl, 2025.
\newblock \url{https://pretty-radio-b75.notion.site/DeepScaleR-Surpassing-O1-Preview-with-a-1-5B-Model-by-Scaling-RL-19681902c1468005bed8ca303013a4e2}.
\newblock Notion Blog.

\bibitem[Mao et~al.(2022)Mao, Wang, Liu, Lin, and Xie]{mao-etal-2022-metaweighting}
Yuren Mao, Zekai Wang, Weiwei Liu, Xuemin Lin, and Pengtao Xie.
\newblock {M}eta{W}eighting: Learning to weight tasks in multi-task learning.
\newblock In Smaranda Muresan, Preslav Nakov, and Aline Villavicencio, editors, \emph{Findings of the Association for Computational Linguistics: ACL 2022}, pages 3436--3448, Dublin, Ireland, May 2022. Association for Computational Linguistics.
\newblock \doi{10.18653/v1/2022.findings-acl.271}.
\newblock \url{https://aclanthology.org/2022.findings-acl.271/}.

\bibitem[Pan et~al.(2025)Pan, Zhang, Wang, Yuan, Peng, and Suhr]{tinyzero}
Jiayi Pan, Junjie Zhang, Xingyao Wang, Lifan Yuan, Hao Peng, and Alane Suhr.
\newblock Tinyzero.
\newblock https://github.com/Jiayi-Pan/TinyZero, 2025.
\newblock Accessed: 2025-01-24.

\bibitem[Parashar et~al.(2025)Parashar, Gui, Li, Ling, Vemuri, Olson, Li, Zhang, Caverlee, Kalathil, et~al.]{parashar2025curriculum}
Shubham Parashar, Shurui Gui, Xiner Li, Hongyi Ling, Sushil Vemuri, Blake Olson, Eric Li, Yu~Zhang, James Caverlee, Dileep Kalathil, et~al.
\newblock Curriculum reinforcement learning from easy to hard tasks improves llm reasoning.
\newblock \emph{arXiv preprint arXiv:2506.06632}, 2025.

\bibitem[Polyak(1963)]{polyak1963gradient}
Boris~Teodorovich Polyak.
\newblock Gradient methods for minimizing functionals.
\newblock \emph{Zhurnal vychislitel'noi matematiki i matematicheskoi fiziki}, 3\penalty0 (4):\penalty0 643--653, 1963.

\bibitem[Qi et~al.(2024)Qi, Chen, Wang, Liu, Zheng, and Wang]{qi2024optimizing}
Zhen Qi, Jiajing Chen, Shuo Wang, Bingying Liu, Hongye Zheng, and Chihang Wang.
\newblock Optimizing multi-task learning for enhanced performance in large language models.
\newblock In \emph{2024 4th International Conference on Electronic Information Engineering and Computer Communication (EIECC)}, pages 1179--1183. IEEE, 2024.

\bibitem[Qwen et~al.(2024)Qwen, Yang, Zhang, Hui, Zheng, Yu, Li, Liu, Huang, Wei, et~al.]{qwen2024qwen2}
A~Yang Qwen, Baosong Yang, B~Zhang, B~Hui, B~Zheng, B~Yu, Chengpeng Li, D~Liu, F~Huang, H~Wei, et~al.
\newblock Qwen2. 5 technical report.
\newblock \emph{arXiv preprint}, 2024.

\bibitem[Sener and Koltun(2018)]{sener2018multi}
Ozan Sener and Vladlen Koltun.
\newblock Multi-task learning as multi-objective optimization.
\newblock \emph{Advances in neural information processing systems}, 31, 2018.

\bibitem[Settles et~al.(2007)Settles, Craven, and Ray]{settles2007multiple}
Burr Settles, Mark Craven, and Soumya Ray.
\newblock Multiple-instance active learning.
\newblock \emph{Advances in neural information processing systems}, 20, 2007.

\bibitem[Shao et~al.(2024)Shao, Wang, Zhu, Xu, Song, Bi, Zhang, Zhang, Li, et~al.]{shao2024deepseekmath}
Zhihong Shao, Peiyi Wang, Qihao Zhu, Runxin Xu, Junxiao Song, Xiao Bi, Haowei Zhang, Mingchuan Zhang, YK~Li, et~al.
\newblock Deepseekmath: Pushing the limits of mathematical reasoning in open language models.
\newblock \emph{arXiv preprint arXiv:2402.03300}, 2024.

\bibitem[Sheng et~al.(2025)Sheng, Zhang, Ye, Wu, Zhang, Zhang, Peng, Lin, and Wu]{sheng2024hybridflow}
Guangming Sheng, Chi Zhang, Zilingfeng Ye, Xibin Wu, Wang Zhang, Ru~Zhang, Yanghua Peng, Haibin Lin, and Chuan Wu.
\newblock Hybridflow: A flexible and efficient rlhf framework.
\newblock In \emph{Proceedings of the Twentieth European Conference on Computer Systems}, pages 1279--1297, 2025.

\bibitem[Shi et~al.(2025)Shi, Wu, Song, Zhou, and Zhao]{shi2025efficient}
Taiwei Shi, Yiyang Wu, Linxin Song, Tianyi Zhou, and Jieyu Zhao.
\newblock Efficient reinforcement finetuning via adaptive curriculum learning.
\newblock \emph{arXiv preprint arXiv:2504.05520}, 2025.

\bibitem[Sutton et~al.(1998)Sutton, Barto, et~al.]{sutton1998reinforcement}
Richard~S Sutton, Andrew~G Barto, et~al.
\newblock \emph{Reinforcement learning: An introduction}, volume~1.
\newblock MIT press Cambridge, 1998.

\bibitem[Team et~al.(2023)]{mosaicml2023introducing}
MosaicML~NLP Team et~al.
\newblock Introducing mpt-7b: A new standard for open-source, commercially usable llms.
\newblock \emph{Retrieved December}, 6:\penalty0 2023, 2023.

\bibitem[Wang et~al.(2024)Wang, Kidambi, Sullivan, Agarwal, Dann, Michi, Gelmi, Li, Gupta, Dubey, Rame, Ferret, Cideron, Hou, Yu, Ahmed, Mehta, Hussenot, Bachem, and Leurent]{wang-etal-2024-conditional}
Kaiwen Wang, Rahul Kidambi, Ryan Sullivan, Alekh Agarwal, Christoph Dann, Andrea Michi, Marco Gelmi, Yunxuan Li, Raghav Gupta, Kumar~Avinava Dubey, Alexandre Rame, Johan Ferret, Geoffrey Cideron, Le~Hou, Hongkun Yu, Amr Ahmed, Aranyak Mehta, Leonard Hussenot, Olivier Bachem, and Edouard Leurent.
\newblock Conditional language policy: A general framework for steerable multi-objective finetuning.
\newblock In Yaser Al-Onaizan, Mohit Bansal, and Yun-Nung Chen, editors, \emph{Findings of the Association for Computational Linguistics: EMNLP 2024}, pages 2153--2186, Miami, Florida, USA, November 2024. Association for Computational Linguistics.
\newblock \doi{10.18653/v1/2024.findings-emnlp.118}.
\newblock \url{https://aclanthology.org/2024.findings-emnlp.118/}.

\bibitem[Wang et~al.(2023{\natexlab{a}})Wang, Chen, Zhou, Duan, Cai, Ma, Cui, Li, Pang, Wang, et~al.]{wang2023aurora}
Rongsheng Wang, Haoming Chen, Ruizhe Zhou, Yaofei Duan, Kunyan Cai, Han Ma, Jiaxi Cui, Jian Li, Patrick Cheong-Iao Pang, Yapeng Wang, et~al.
\newblock Aurora: Activating chinese chat capability for mixtral-8x7b sparse mixture-of-experts through instruction-tuning.
\newblock \emph{arXiv preprint arXiv:2312.14557}, 2023{\natexlab{a}}.

\bibitem[Wang et~al.(2023{\natexlab{b}})Wang, Ivison, Dasigi, Hessel, Khot, Chandu, Wadden, MacMillan, Smith, Beltagy, et~al.]{wang2023far}
Yizhong Wang, Hamish Ivison, Pradeep Dasigi, Jack Hessel, Tushar Khot, Khyathi Chandu, David Wadden, Kelsey MacMillan, Noah~A Smith, Iz~Beltagy, et~al.
\newblock How far can camels go? exploring the state of instruction tuning on open resources.
\newblock \emph{Advances in Neural Information Processing Systems}, 36:\penalty0 74764--74786, 2023{\natexlab{b}}.

\bibitem[Wang et~al.(2025)Wang, Cui, Li, Wan, and Zhao]{wang2025dump}
Zhenting Wang, Guofeng Cui, Yu-Jhe Li, Kun Wan, and Wentian Zhao.
\newblock Dump: Automated distribution-level curriculum learning for rl-based llm post-training.
\newblock \emph{arXiv preprint arXiv:2504.09710}, 2025.

\bibitem[Yang et~al.(2024)Yang, Pan, Luo, Qiu, Zhong, Yu, and Chen]{yang2024rewards}
Rui Yang, Xiaoman Pan, Feng Luo, Shuang Qiu, Han Zhong, Dong Yu, and Jianshu Chen.
\newblock Rewards-in-context: Multi-objective alignment of foundation models with dynamic preference adjustment.
\newblock In \emph{International Conference on Machine Learning}, pages 56276--56297. PMLR, 2024.

\bibitem[Yu et~al.(2025)Yu, Zhang, Zhu, Yuan, Zuo, Yue, Dai, Fan, Liu, Liu, et~al.]{yu2025dapo}
Qiying Yu, Zheng Zhang, Ruofei Zhu, Yufeng Yuan, Xiaochen Zuo, Yu~Yue, Weinan Dai, Tiantian Fan, Gaohong Liu, Lingjun Liu, et~al.
\newblock Dapo: An open-source llm reinforcement learning system at scale.
\newblock \emph{arXiv preprint arXiv:2503.14476}, 2025.

\bibitem[Yu et~al.(2020)Yu, Kumar, Gupta, Levine, Hausman, and Finn]{yu2020gradient}
Tianhe Yu, Saurabh Kumar, Abhishek Gupta, Sergey Levine, Karol Hausman, and Chelsea Finn.
\newblock Gradient surgery for multi-task learning.
\newblock \emph{Advances in neural information processing systems}, 33:\penalty0 5824--5836, 2020.

\bibitem[Zhang et~al.(2025)Zhang, Arora, Mei, and Zanette]{zhang2025speed}
Ruiqi Zhang, Daman Arora, Song Mei, and Andrea Zanette.
\newblock Speed-rl: Faster training of reasoning models via online curriculum learning.
\newblock \emph{arXiv preprint arXiv:2506.09016}, 2025.

\bibitem[Zhu et~al.(2025)Zhu, Dong, Qu, Ruan, Chen, and Cheng]{zhu2025dynamic}
Tong Zhu, Daize Dong, Xiaoye Qu, Jiacheng Ruan, Wenliang Chen, and Yu~Cheng.
\newblock Dynamic data mixing maximizes instruction tuning for mixture-of-experts.
\newblock In \emph{Proceedings of the 2025 Conference of the Nations of the Americas Chapter of the Association for Computational Linguistics: Human Language Technologies (Volume 1: Long Papers)}, pages 1663--1677, 2025.

\bibitem[Łojasiewicz(1963)]{Lojasiewicz1963}
Stanisław Łojasiewicz.
\newblock Une propriété topologique des sous-ensembles analytiques réels.
\newblock In \emph{Les Équations aux Dérivées Partielles}, Colloques Internationaux du CNRS, pages 87--89. Éditions du CNRS, 1963.

\end{thebibliography}

\clearpage
\appendix

\section{Prior Works on Multi-Task LLMs}\label{sec:works}

Prior works on multi-task LLMs have approached the problem from task weighting and data mixture design~\citep{mao-etal-2022-metaweighting,gong2024coba,wang-etal-2024-conditional,yang2024rewards,feng2024mixture,elbakary2025mira,liang2025modomodo}. Typically, they adaptively adjust task contributions or update directions to balance convergence across tasks.

Curriculum-based methods have also emerged in the context of LLMs, which schedule tasks adaptively based on a certain notion of learning progress or difficulty~\citep{chen2025self,shi2025efficient,parashar2025curriculum}. For instance, it may suggest training on progressively harder reasoning instances. While these approaches are usually proposed under a single task and manipulate at a sample level, their natural extension can be applied to the multi-task settings.

In contrast, our study focuses on a fundamental issue: the imbalance of gradient magnitudes across tasks during RL post-training. While similar concerns have been observed in the broader field such as vision~\citep{chen2018gradnorm}, this issue has received less attention in the context of LLMs.

\section{Gradient Magnitude and Learning Gains Through the Lens of Convex Analysis}\label{sec:convex-analysis}

In this section, we study the relationship between gradient magnitude and the learning gain through the lens of convex analysis. Specifically, we will explain why gradient magnitude (squared gradient norm) is expected to be proportional to learning gains \emph{within a task} but not \emph{across tasks} from a theoretical perspective.

Let's think about a single task and thus a single objective optimization problem. Denote the objective function as $f$.
To facilitate the discussion, we consider two common properties of convex functions: smoothness and Polyak-\L{}ojasiewicz condition (a relaxation of strong convexity).
\begin{definition}[Smoothness]
A function $f : \mathbb{R}^n \to \mathbb{R}$ is said to be \emph{$\beta$-smooth} if for all $x, y \in \mathbb{R}^n$,
\[
    f(y) \leq f(x) + \nabla f(x)^\top (y - x) + \frac{\beta}{2}\|y - x\|_2^2.
\]
\end{definition}
It is known that smoothness relates the gradient magnitude to the suboptimality gap~\citep{hazan2016introduction}:
\begin{equation}\label{eq:smooth}
f(x) - \min_{x^\star} f(x^\star) \geq \frac{1}{2\beta} \|\nabla f(x)\|_2^2.
\end{equation}
Another commonly used property is the Polyak-\L{}ojasiewicz condition:
\begin{definition}[Polyak-\L{}ojasiewicz]\citep{Lojasiewicz1963,polyak1963gradient}
A function $f : \mathbb{R}^n \to \mathbb{R}$ is said to be \emph{$\mu$-Polyak--\L{}ojasiewicz} (\emph{$\mu$-PL}) if for all $x \in \mathbb{R}^n$, we have
\begin{equation}\label{eq:pl}
    \frac{1}{2} \|\nabla f(x)\|_2^2 \geq \mu \cdot \big(f(x) - \min_{x^\star} f(x^\star)\big).
\end{equation}
\end{definition}
PL condition generalizes strong convexity~\citep{hazan2016introduction}. By definition, it relates the gradient magnitude to the suboptimality gap in a reversed direction. By \eqref{eq:smooth} and \eqref{eq:pl}, one can establish the following ratio bound:
\[
2 \mu \leq \frac{\|\nabla f(x)\|_2^2}{f(x) - \min_{x^\star} f(x^\star)} \leq 2\beta 
\]
This inequality suggests that the squared gradient norm should be proportional to the suboptimality gap. However, estimating the gap in practice is difficult since it requires access to the global minimizer $\min_{x^\star} f(x^\star)$. A useful observation is that performance improvements (i.e., learning gain) during training often follow an exponential decay: once the suboptimality gap becomes small, further gains diminish. Thus, the learning gain serves as a practical proxy for the suboptimality gap, implying that \emph{the magnitude of squared gradient norm should be proportional to learning gain}. This is the fundamentals of certain existing works that leverage gradient magnitude to estimate learning gains.

However, from a theoretical perspective, this proportionality holds only within a single task. When comparing across tasks, the relationship breaks down because each task may have different smoothness $\beta$ and PL constant $\mu$. This heterogeneity explains the mismatch between gradient magnitude and learning gain across tasks observed in \Cref{fig:intro}.

\section{Experimental Details}\label{sec:exp-details}

This section details the experiment setup. All implementations are built upon the verl framework \citep{sheng2024hybridflow}. 

\lstdefinestyle{mystyle}{
basicstyle=\ttfamily,          %
breaklines=true,               %
keywordstyle=\color{orange}\bfseries, %
}
\lstset{style=mystyle}

\paragraph{Prompt Template.}
We use the same prompt template for all tasks and all models. Given a \texttt{question}, the prompt template is as follows:
\begin{lstlisting}
<|system|>
You are a helpful assistant. The user will ask you a question and you as the assistant solve it. The assistant first thinks how to solve the task step by step and then provides the user with the final answer. The final answer must be enclosed within <answer>...</answer> tag, which should appear at the end of your reply.
<|user|>
{question}
<|Assistant|>
\end{lstlisting}

\paragraph{Reward Details.}
We use a rule-based reward function.
A reward of 0.0 is assigned to a response if the format is wrong (e.g., answer tag \texttt{<answer>...</answer>} is not found). Otherwise, if the format is correct but answer is wrong, we assign a reward of 0.1. If both are correct, we assign a reward of 1.0.

\paragraph{Model and Dataset Details.}
We have listed the models and datasets we used in \Cref{sec:imb}.
For each task/dataset, the data is split into training and test sets (with MATH500 serving as the test set for MATH). 

\begin{figure*}[htb]
\centering
\includegraphics[width=\linewidth]{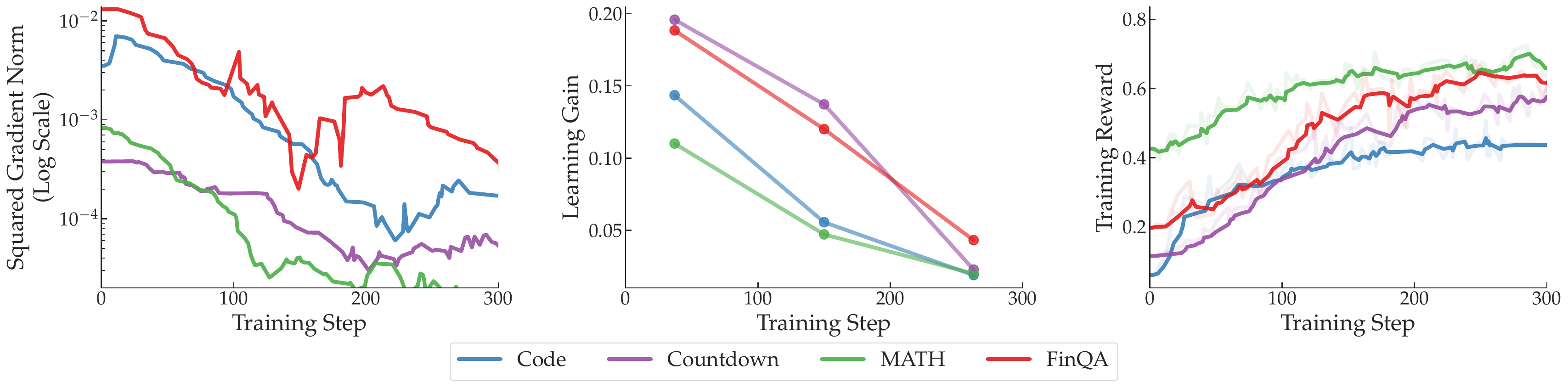}

\vspace{0.7em}

\includegraphics[width=\linewidth]{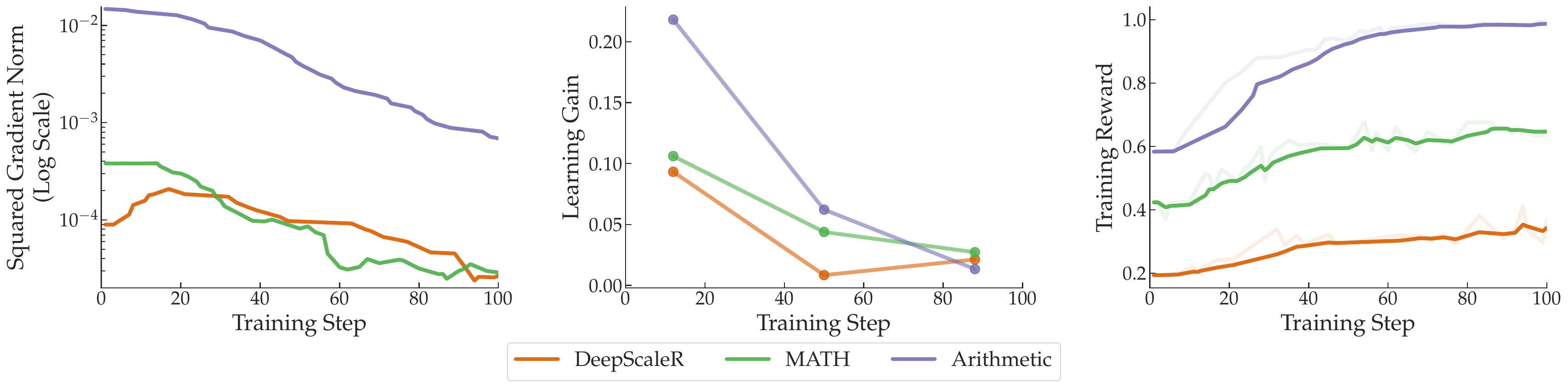}

\vspace{0.7em} 
\includegraphics[width=\linewidth]{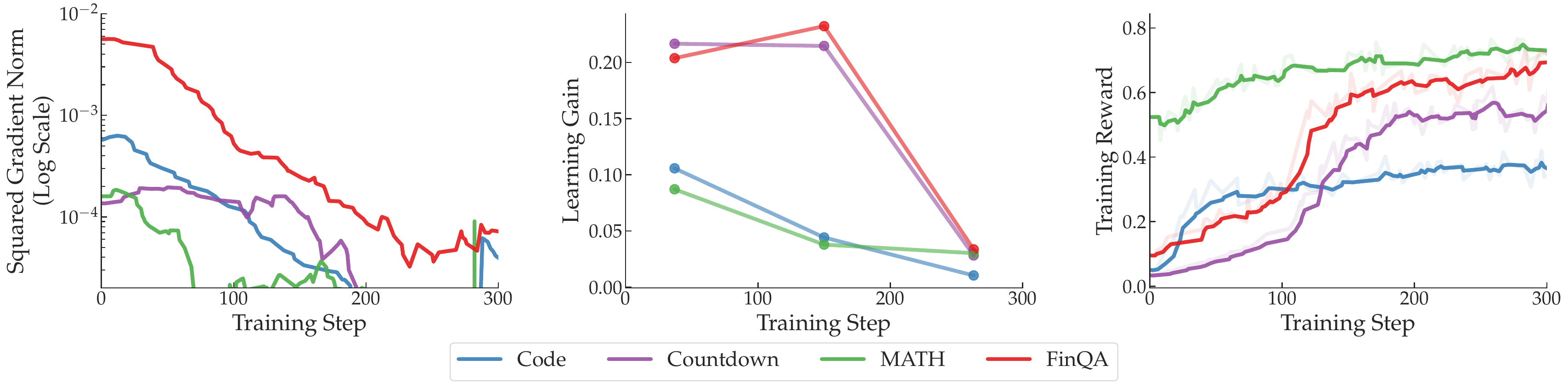}

\vspace{0.7em}

\includegraphics[width=\linewidth]{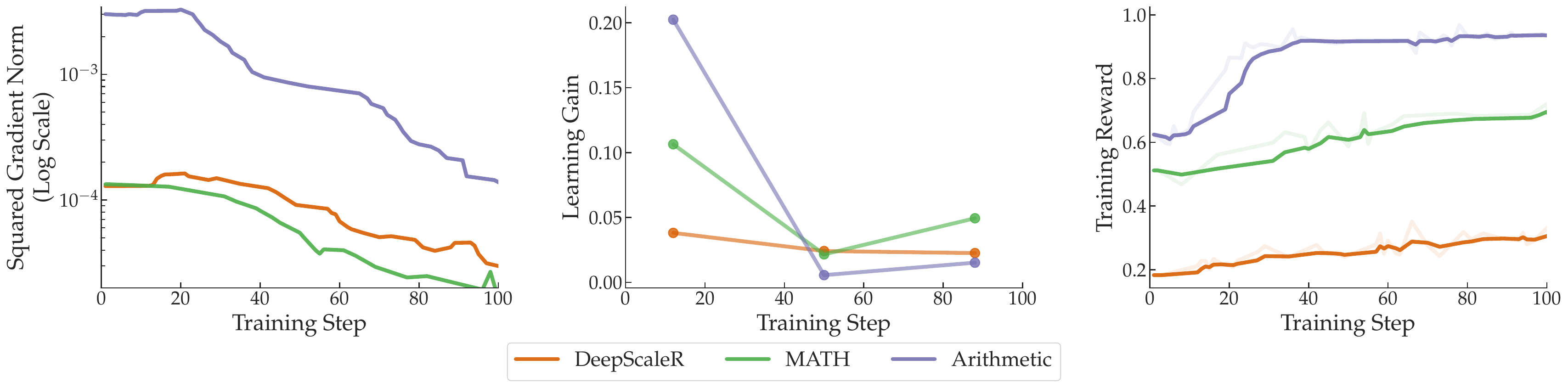}
\vspace{-0.3em}

\caption{
    \textbf{Replication of gradient imbalance across models.} 
    Same setup as Figure~\ref{fig:intro} but for Qwen-3B (top two rows) and Llama-3B (bottom two rows). The imbalance pattern persists.
}

\label{fig:intro-additional}
\end{figure*}

\begin{figure*}[htb]
\centering
\includegraphics[width=0.98\linewidth]{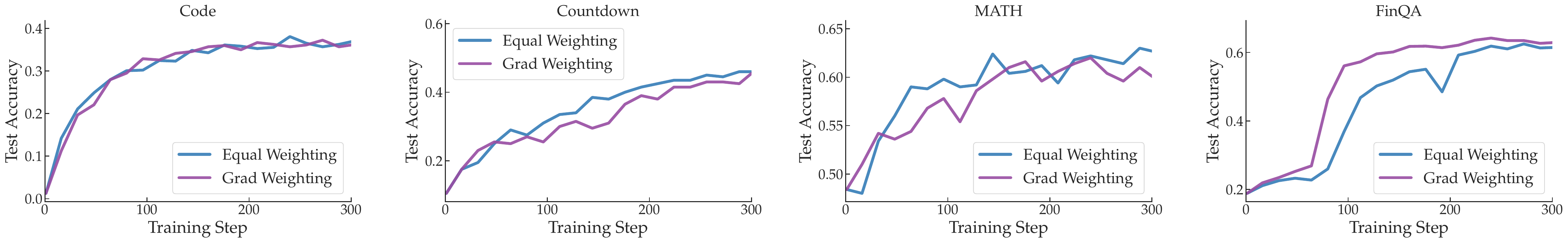}
\vspace{0.2cm}
\includegraphics[width=0.98\linewidth]{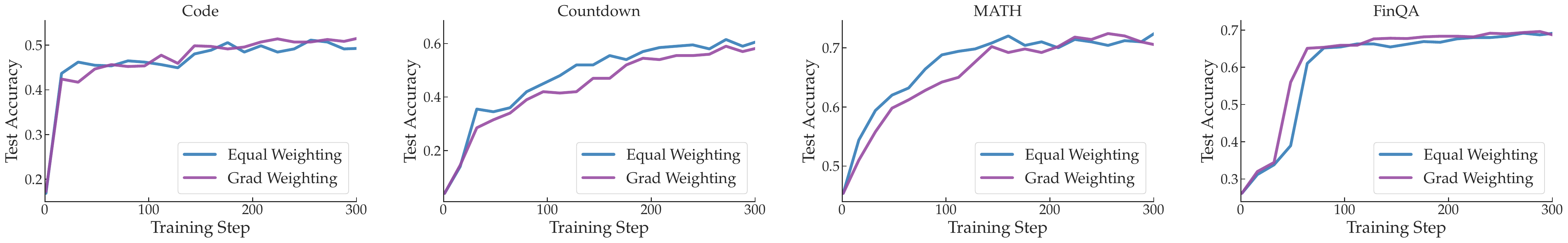}
\vspace{0.2cm}
\includegraphics[width=0.98\linewidth]{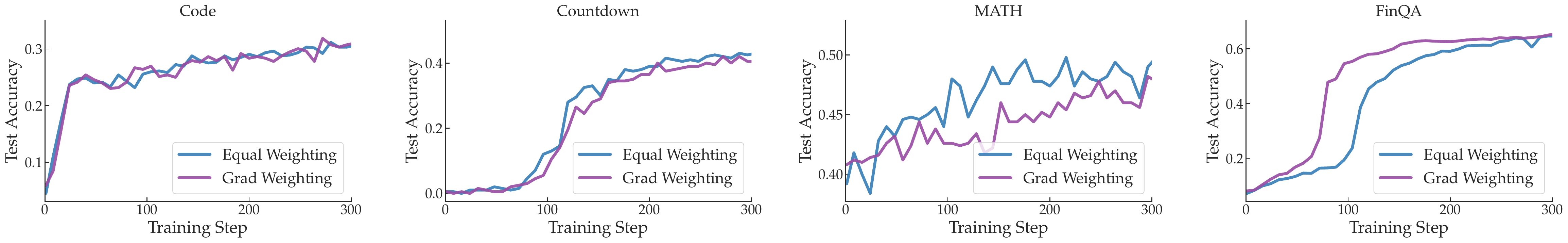}
\vspace{0.2cm}
\includegraphics[width=0.8\linewidth]{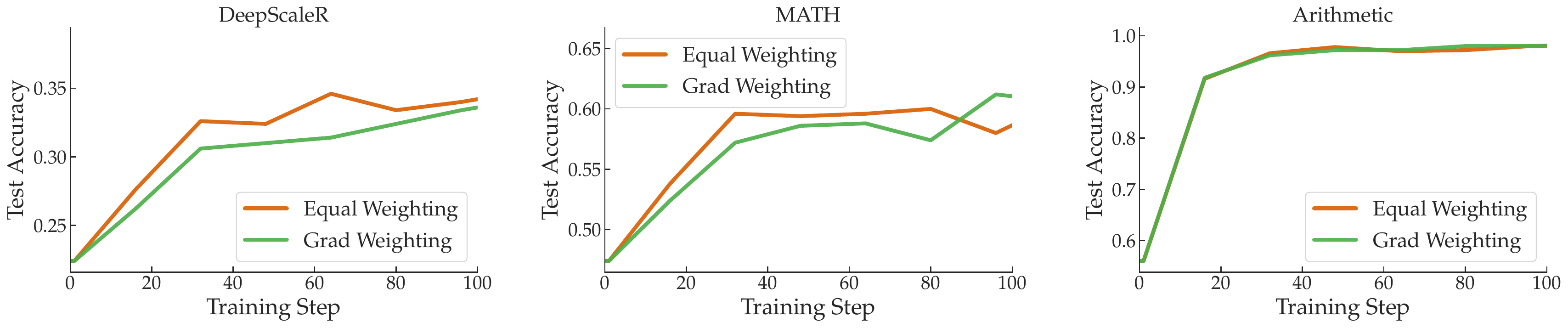}
\vspace{0.2cm}
\includegraphics[width=0.8\linewidth]{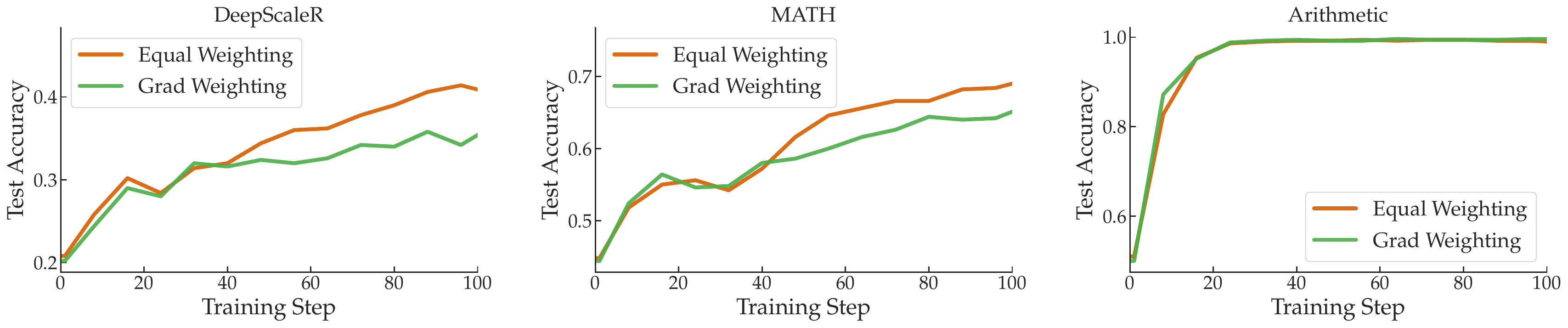}
\vspace{0.2cm}
\includegraphics[width=0.8\linewidth]{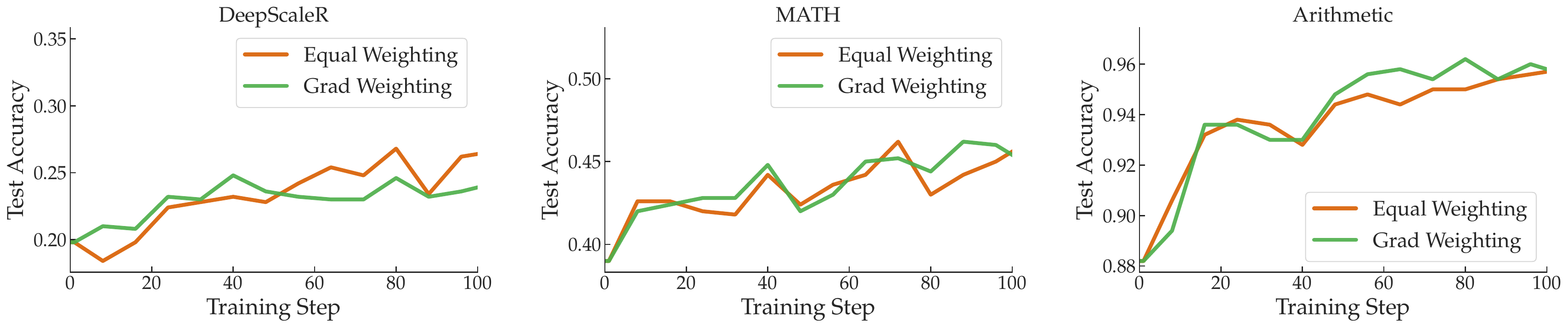}
\caption{\textbf{Task-level accuracy under uniform vs. gradient-proportional sampling} (as reported in \Cref{tab:4t}). Top three rows present multi-domain tasks, and bottom three present single-domain tasks. Within each block, results are shown for Qwen-3B, Qwen-7B, and Llama-3B from top to bottom.}
\label{fig:training-curves}
\end{figure*}

\paragraph{Plotting details.}
In \Cref{fig:intro}, we smooth the squared gradient norm (left plots) via exponential moving average with coefficient 0.9 for better illustration; for the learning gains (middle plots), we smooth rewards over $s=75$ steps for multi-domain setting and $s=25$ for single-domain setting; the reward plots on the right are smoothed with coefficient 0.7 with the raw data plotted as the lighter curves. Same also applies to \Cref{fig:intro-additional} and \Cref{fig:grad-ablation}.
\Cref{fig:adv_vs_grad_norm} is plotted by uniformly picking points throughout training and computing the absolute advantage and gradient norm at those points (without smoothing). \Cref{fig:response-length} is plotted with smoothing factor 0.5 with the raw data plotted as the lighter curves. 
\Cref{fig:training-curves} is plotted without smoothing.

\paragraph{Hyperparameters.}
Hyperparameters are listed in \Cref{tab:hyperparam} and are kept consistent across all training runs. Training was performed via GRPO on A100 GPUs. For both multi-domain and single-domain settings, we train the model until observed convergence (300 steps for the former and 100 steps for the latter).
\begin{table}[h]
\centering
\begin{tabular}{lc}
    \toprule
    {Parameter} & {Value} \\
    \midrule
    batch size & 128 \\
    rollouts per prompt & 16 \\
    max prompt length & 2048 \\
    max response length & 4096 \\
    learning rate & 5e-7 \\
    KL coefficient & 1e-3 \\
    entropy coefficient & 1e-3 \\
    grad clip & 1.0 \\
    clip ratio & 0.2 \\
    \bottomrule
\end{tabular}
\caption{\textbf{Training hyperparameters.} All experiment runs are performed with the same hyperparameters.}
\label{tab:hyperparam}
\end{table}

\subsection{Gradient Estimation}\label{sec:grad-est}

In this section, we explain how we estimate the gradient magnitude in the experiments (e.g., in \Cref{fig:intro}). To obtain a faithful estimator, we cannot simply take the norm of a batch gradient returned by back-propagation. To see why, let $g$ denote the ground truth gradient we want to estimate:
\[
g \;=\; \nabla_\theta J(\theta)
\;=\; \mathbb{E}\!\left[\,X\,\right],
\]
\[
\text{where}\;\; X \;\coloneqq\; \nabla_\theta \log \pi_\theta(a \mid s)\,A^\pi(s,a).
\]
Given a batch of size $B$, the naive plug-in estimator is
\[
\hat g \;=\; \frac{1}{B}\sum_{i=1}^B X_i,
\qquad X_1,\ldots,X_B \stackrel{\text{i.i.d.}}{\sim} X.
\]
While $\hat g$ is an unbiased estimator of $g$, its \emph{squared norm} is a biased estimator of $\|g\|^2$. To see this, using the identity:
\[
\mathbb{E}\!\left[\|\hat g\|^2\right]
\;=\; \bigl\|\mathbb{E}[\hat g]\bigr\|^2 \;+\; \operatorname{Tr}\bigl(\operatorname{Cov}(\hat g)\bigr),
\]
we obtain
\begin{align*}
\mathbb{E}\!\left[\|\hat g\|^2\right]
&\;=\; \|g\|^2 \;+\; \operatorname{Tr}\bigl(\operatorname{Cov}(\hat g)\bigr) \\
&\;=\; \|g\|^2 \;+\; \frac{1}{B}\,\operatorname{Tr}\bigl(\Sigma\bigr),
\end{align*}
where $\Sigma \coloneqq \operatorname{Cov}(X)$. Thus, the naive estimator \(\|\hat g\|^2\) is \emph{off from the target} \(\|g\|^2\) by the variance term \(\operatorname{Tr}(\operatorname{Cov}(\hat g)) = \frac{1}{B}\operatorname{Tr}(\Sigma)\). This term is large considering the huge number of parameters in LLMs. 

To remove this bias, we form two independent estimates by splitting the batch:
\[
\hat g_1 \;=\; \frac{2}{B}\sum_{i=1}^{B/2} X_i,
\qquad
\hat g_2 \;=\; \frac{2}{B}\sum_{i=B/2+1}^{B} X_i,
\]
and use the cross product
\[
\widehat{\|g\|^2} \;=\; \langle \hat g_1,\hat g_2\rangle.
\]
Since $\hat g_1$ and $\hat g_2$ are independent and $\mathbb{E}[\hat g_1]=\mathbb{E}[\hat g_2]=g$, we have the unbiasedness:
\[
\mathbb{E}\!\left[\,\langle \hat g_1,\hat g_2\rangle\,\right]
= \bigl\langle \mathbb{E}[\hat g_1],\,\mathbb{E}[\hat g_2]\bigr\rangle
= \langle g, g\rangle
= \|g\|^2.
\]
Hence, such cross-product estimator provides a faithful (and importantly, unbiased) estimate of the squared gradient norm. In practice, we track the squared gradient norm via the above estimator per step and keep an exponential moving average (with coefficient $0.95$) of the tracked values as the final estimate. Furthermore, when we want to estimate the (unsquared) gradient norm, we simply take the square root of the cross-product estimator above, truncated at zero:
\[
\hat{\|g\|} \;=\; \sqrt{\max\left(\widehat{\|g\|^2}, 0\right)}.
\]
Here we truncate at zero to avoid negative values inside the square root since the cross-product estimator is not guaranteed to be non-negative. To reduce the overall computational cost of the cross product, we only compute the gradient of the last attention block as a proxy for the entire model. An ablation study on this choice is provided in \Cref{sec:ablation-study}.

\section{Additional Experiment Results}\label{sec:additional-exp-results}

\subsection{Ablation Study}\label{sec:ablation-study}

\paragraph{Gradient-Proportional Sampling.}
We perform a sweep of temperature for gradient-proportional sampling on all models for multi-domain tasks and report the results in \Cref{tab:ablation-study}. For $\eta \geq 0.1$, we find the resulting weights are nearly identical to uniform sampling, so meaningful weight differences appear only when $\eta < 0.1$, which is why we mainly focus on the range below $\eta=0.1$. However, as the temperature decreases, we consistently observe degraded average performance, indicating that over-weighting large-gradient tasks is detrimental. We fix $\eta=0.01$ to report results in \Cref{tab:4t} in the main text across all models and settings.

\begin{table*}[t]
\small
\centering
\begin{tabular}{lccccccc}
    \toprule
    \multirow{2}{*}{{Model}} 
    & \multirow{2}{*}{\makecell{Grad-Prop\\Sampling?}}
    & \multirow{2}{*}{{Temperature}} 
    & \multicolumn{5}{c}{{Multi-domain}} 
    \\
    \cmidrule{4-8}
    & & & {Code} & {Countdown} & {MATH} & {FinQA} & \textit{{Avg}}
    \\
    \midrule
    \multirow{4}{*}{Qwen-3B} & \ding{55} & - & 35.67 & {50.00} & {63.20} & 60.60 & \textbf{52.37} \\
    & \ding{51} & 0.1 & 35.53 & 48.50 & 60.00 & 61.59 & 51.41 \\
    & \ding{51} & 0.01 & 35.67 & 42.50 & 61.00 & 62.67 & 50.46 \\
    & \ding{51} & 0.001 & {38.06} & 37.50 & 62.60 & {62.85} & 50.25 \\
    \midrule
    \multirow{3}{*}{Lllama-3B} & \ding{55} & - &  {30.34} & {42.50} & {49.00} & 64.65 & \textbf{46.62} \\
    & \ding{51} & 0.1 & 28.51 & 40.00 & 44.20 & 62.04 & 43.69 \\
    & \ding{51} & 0.01 & 30.76 & 40.50 & 48.20 & {65.01} & 46.12 \\
    & \ding{51} & 0.001 & 29.92 & 39.50 & 46.80 & 64.47 & 45.17 \\
    \midrule
    \multirow{2}{*}{Qwen-7B} & \ding{55} & - &  49.16 & {59.00} & {71.00} & {68.71} & \textbf{61.97} \\
    & \ding{51} & 0.1 & {50.98} & 56.50 & 70.20 & 68.44 & 61.53 \\
    & \ding{51} & 0.01 & 50.70 & 53.00 & 69.60 & 68.62 & 60.48 \\
    & \ding{51} & 0.001 & 52.25 & 52.00 & 71.00 & 69.52 & 61.19 \\
    \bottomrule
\end{tabular}

\caption{\textbf{Sweep of gradient-proportional sampling temperature across models.} For temperature $\eta \geq 0.1$, we find the resulting weights are nearly identical to uniform sampling, so we mainly focus on the range below $0.1$. As the temperature decreases, we consistently observe degraded average performance, indicating that over-weighting large-gradient tasks is detrimental. This further demonstrates the pitfalls of naively treating gradient magnitude as a proxy for learning gains.}

\label{tab:ablation-study}
\end{table*}

\begin{figure*}[htb]
\centering
\includegraphics[width=0.32\linewidth]{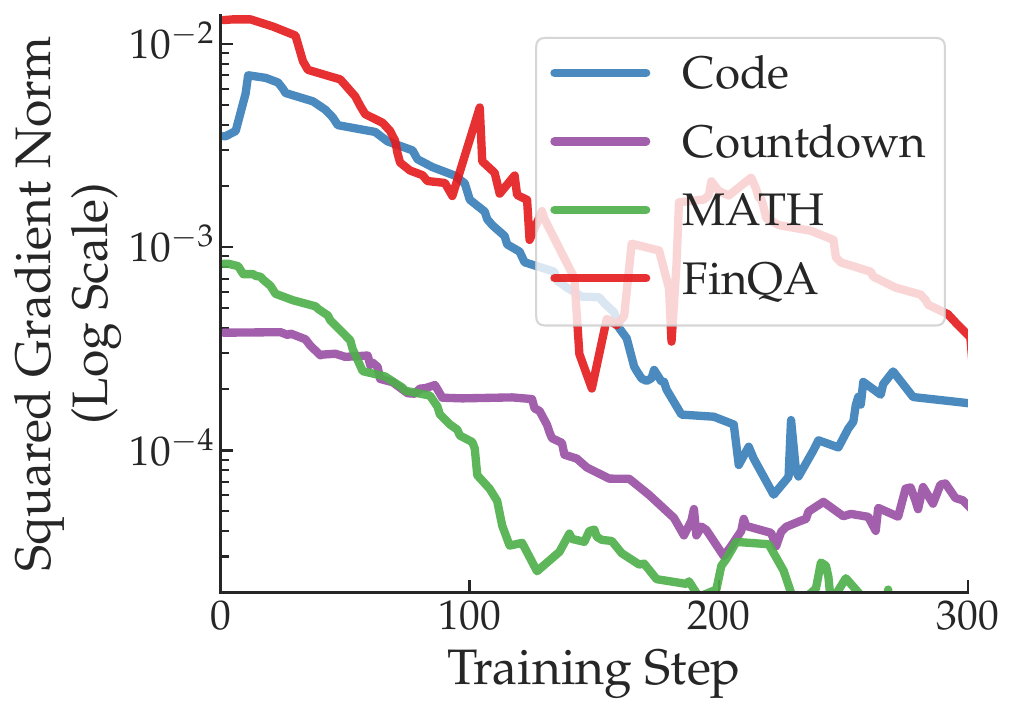}
\hfill
\includegraphics[width=0.32\linewidth]{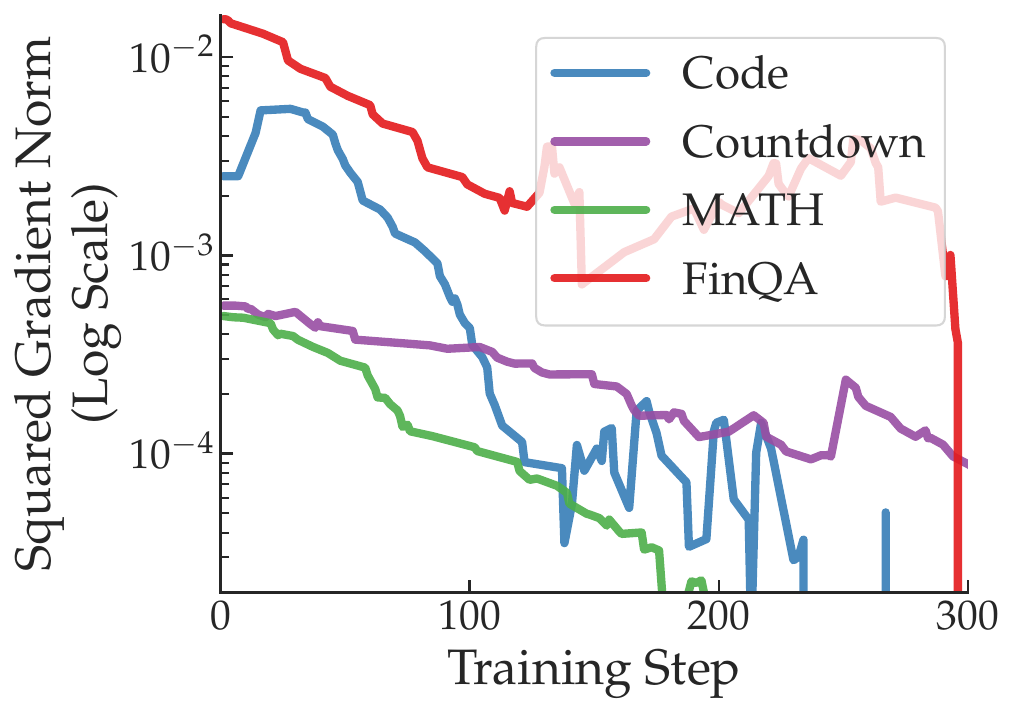}
\hfill
\includegraphics[width=0.32\linewidth]{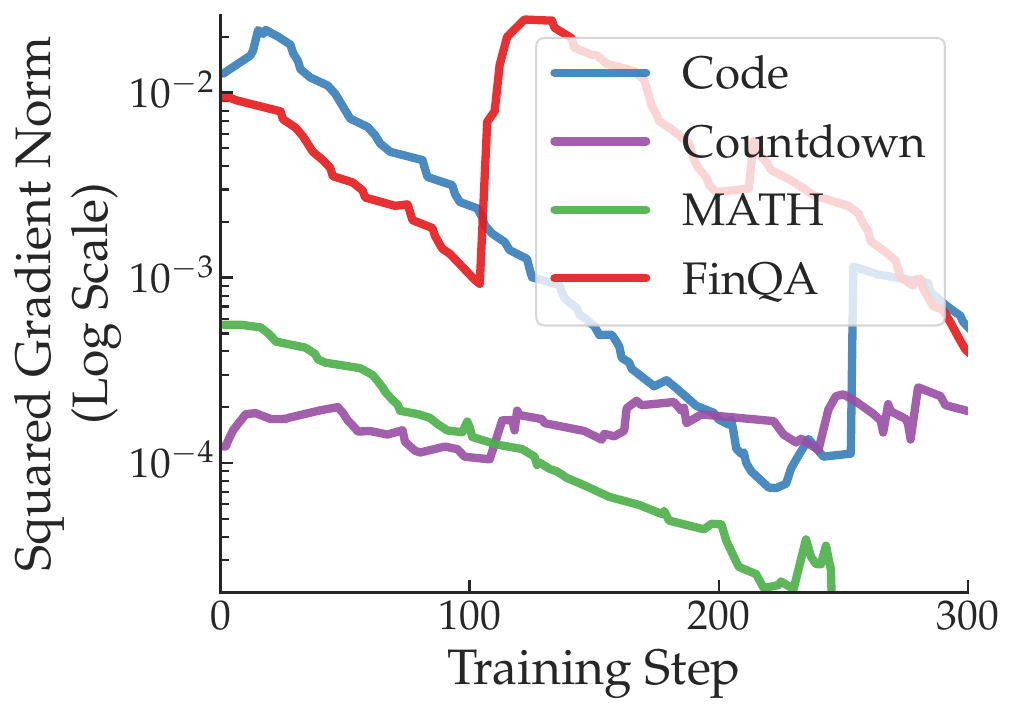}
\caption{\textbf{Gradient norm of different attention blocks for Qwen-3B model on multi-domain tasks.} There are a total of 36 attention blocks in the model. From left to right are the gradients of the last (36th) attention block, middle (18th) attention block, and the first attention block. }
\label{fig:grad-ablation}
\end{figure*}

\paragraph{Gradients of Different Attention Blocks.}
To reduce computational cost, we approximate the full-model gradient by computing it only on the last attention block, as described in \Cref{sec:grad-est}. To validate this choice, we examine the Qwen-3B model by computing gradients across other layers (first, middle, last) and reporting them in \Cref{fig:grad-ablation}. The results show that gradients across layers are of comparable scale and the dominance pattern remains consistent across layers: Code and FinQA have the largest gradients, while MATH has the smallest. Notably, earlier blocks (first and middle blocks) exhibit higher variance. These confirm that the last attention block serves as a reliable proxy.

\subsection{Supplementary Plots}\label{sec:additional-plots}

In this section, we supplement additional plots of other models and statistics for completeness. The observations are consistent with the main text.

\begin{figure*}[htb]
\centering
\includegraphics[width=0.32\linewidth]{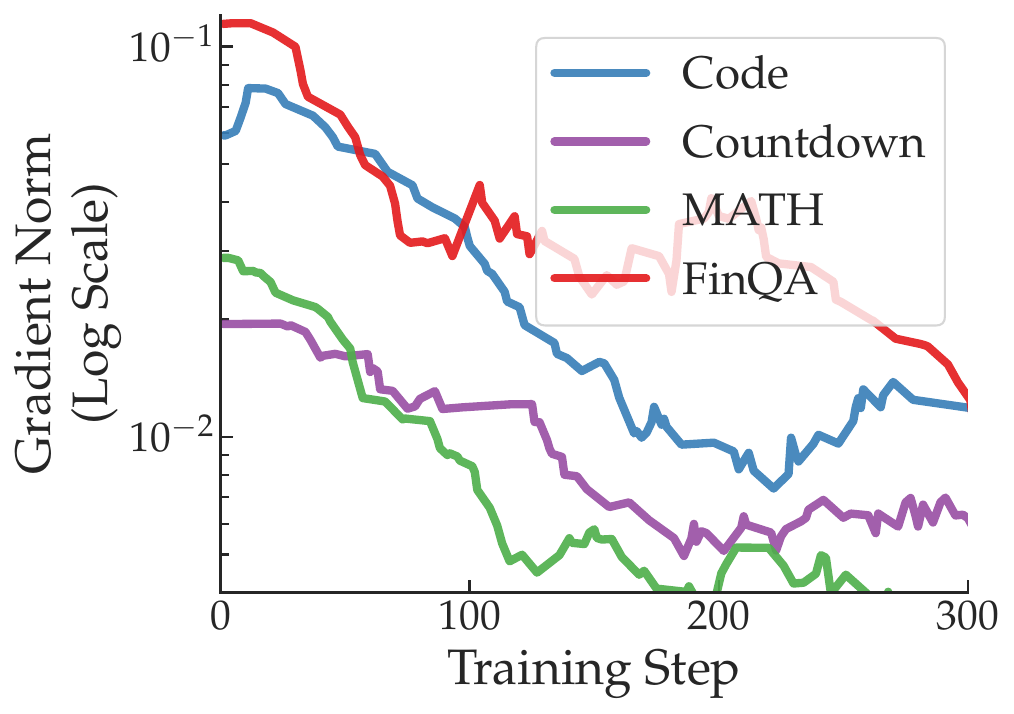}
\hfill
\includegraphics[width=0.32\linewidth]{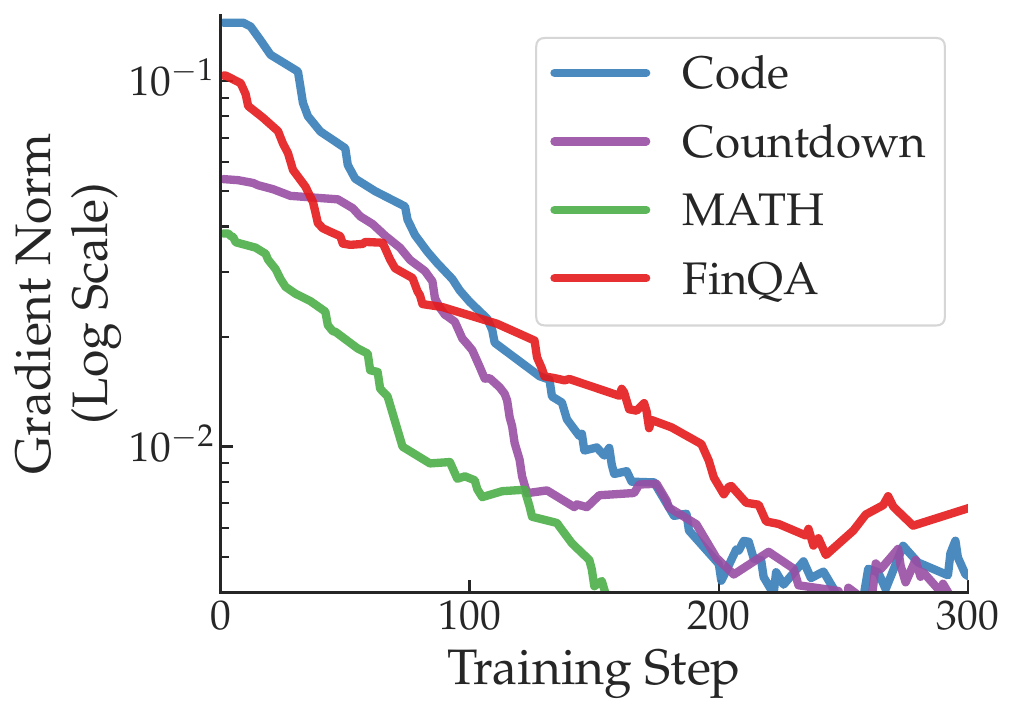}
\hfill
\includegraphics[width=0.32\linewidth]{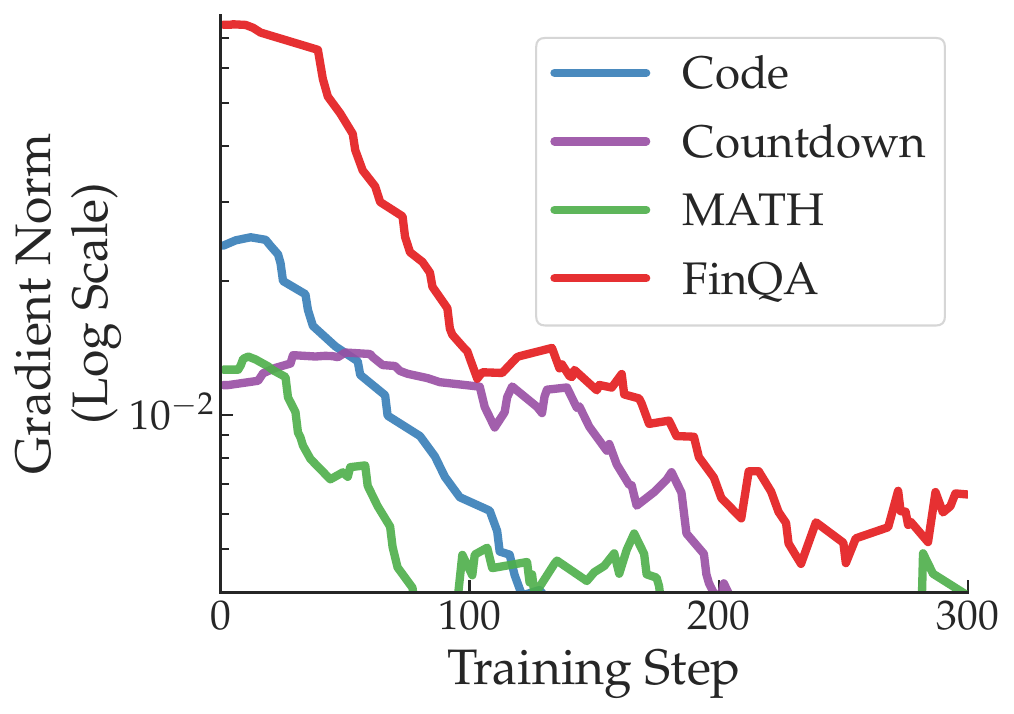}
\includegraphics[width=0.32\linewidth]{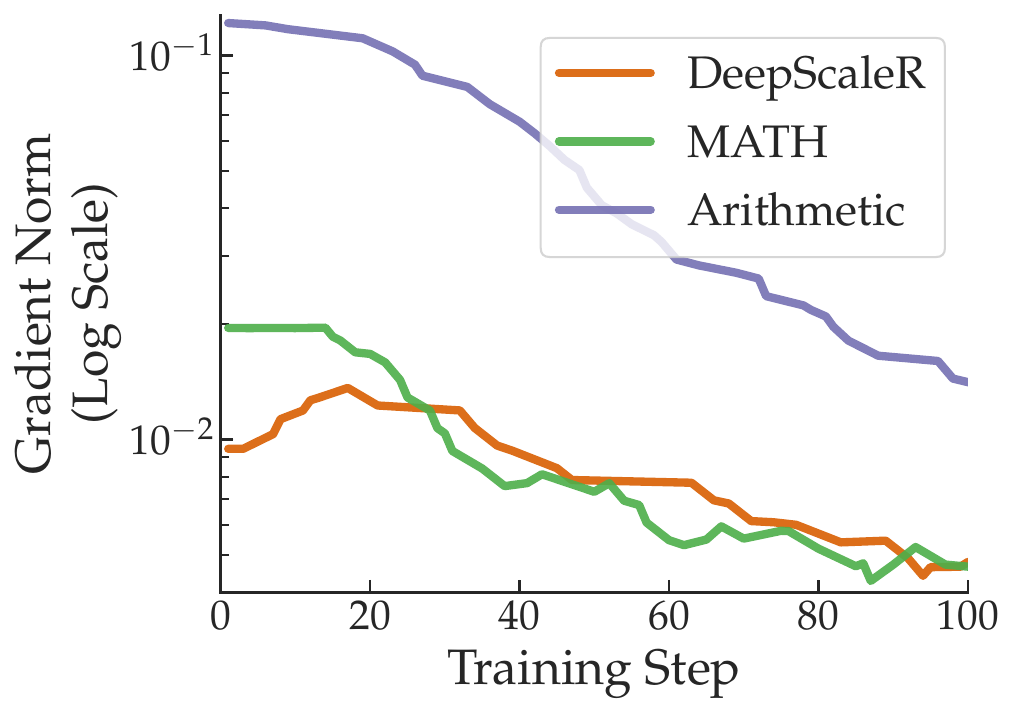}
\hfill
\includegraphics[width=0.32\linewidth]{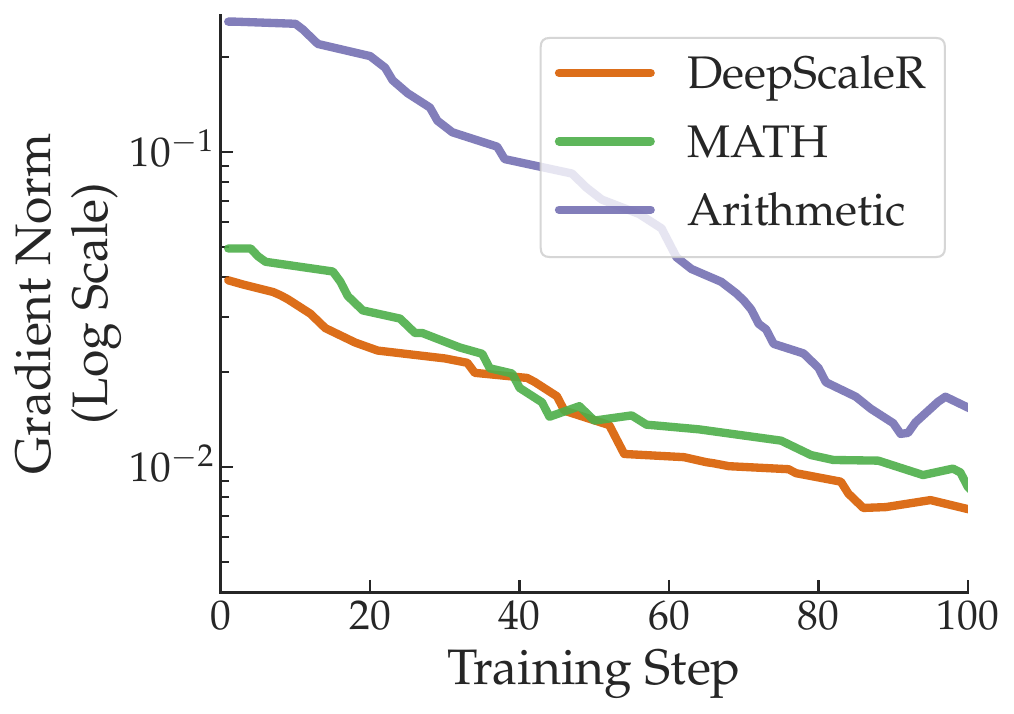}
\hfill
\includegraphics[width=0.32\linewidth]{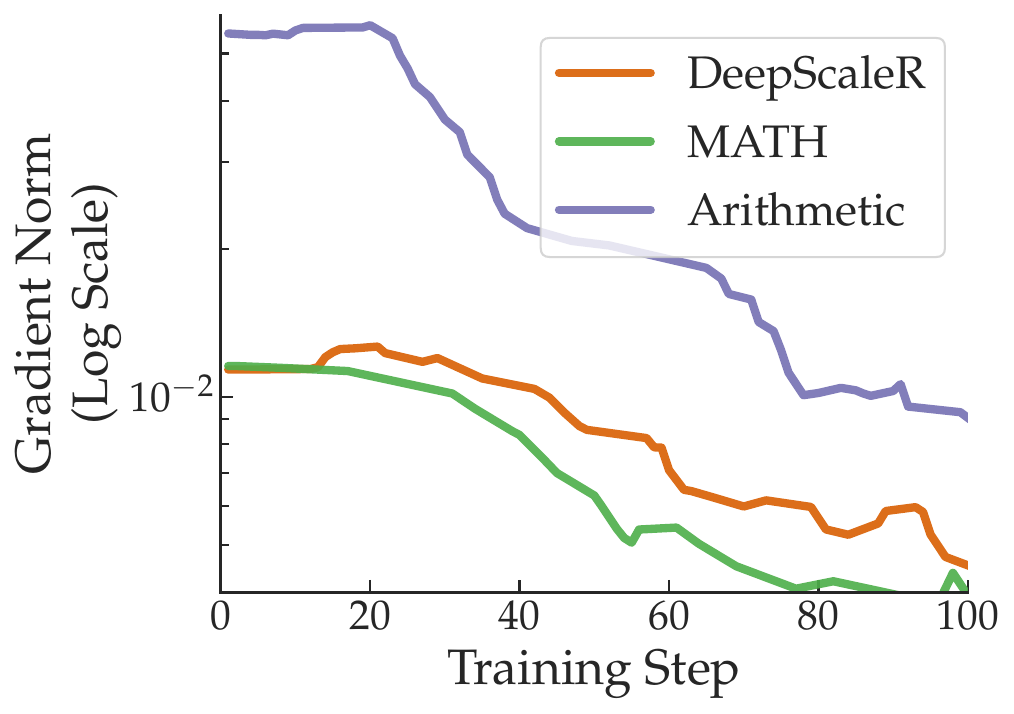}
\caption{\textbf{(Unsquared) Gradient norm across models.} From left to right are Qwen-3B, Qwen-7B, and Llama-3B. Top row presents multi-domain tasks, and bottom row presents single-domain tasks. The pattern is identical to that of the squared norm in \Cref{fig:intro} and \Cref{fig:intro-additional}.}
\label{fig:unsq_grad_norm}
\end{figure*}

\end{document}